%% file: acl_latex.tex
\documentclass[11pt]{article}

\usepackage[preprint]{acl}

\usepackage{times}
\usepackage{latexsym}

\usepackage[T1]{fontenc}

\usepackage[utf8]{inputenc}

\usepackage{microtype}

\usepackage{inconsolata}

\usepackage{wrapfig}
\usepackage{subcaption}
\usepackage{graphicx}
\usepackage{enumitem}
\usepackage{placeins}
\usepackage{amsmath}   
\usepackage{amssymb}   
\usepackage{bm}        
\usepackage{booktabs}   
\usepackage{soul} 
\usepackage{multirow} 
\usepackage{pifont}
\usepackage{longtable}
\usepackage{tabularx}

\usepackage{svg}
\svgpath{{../img/}}

\usepackage{pgfmath}              
\usepackage[table]{xcolor}        
\usepackage{booktabs}             

\usepackage[hang,flushmargin]{footmisc}

\definecolor{LightRed}{HTML}{F0A9A3}
\definecolor{LightYellow}{HTML}{ffef95}
\sethlcolor{LightRed}

\expandafter\def\expandafter\normalsize\expandafter{%
    \normalsize%
    \setlength\abovedisplayskip{4.5pt}%
    \setlength\belowdisplayskip{4.5pt}%
    \setlength\abovedisplayshortskip{3pt}%
    \setlength\belowdisplayshortskip{3pt}%
}

\definecolor{gradBad}{HTML}{b1cbd5}   
\definecolor{gradMid}{HTML}{fefefe}   
\definecolor{gradGood}{HTML}{ecc4ae}  

\makeatletter
\newcommand{\@gradmix}[3]{
  \pgfmathtruncatemacro{\@gpct}{min(100,max(0,(0#3-#1)/(#2-#1)*100))}%
  \ifnum\@gpct<50
    \pgfmathtruncatemacro{\@gloc}{\@gpct*2}%
    \edef\gradcellcmd{\noexpand\cellcolor{gradMid!\@gloc!gradBad}}%
  \else
    \pgfmathtruncatemacro{\@gloc}{(\@gpct-50)*2}%
    \edef\gradcellcmd{\noexpand\cellcolor{gradGood!\@gloc!gradMid}}%
  \fi
}
\newcommand{\gradcell}[3]{\@gradmix{#1}{#2}{#3}\gradcellcmd #3}
\newcommand{\gradcellbf}[3]{\@gradmix{#1}{#2}{#3}\gradcellcmd \textbf{#3}}
\makeatother

\newcommand{\cJSD}[1]{\gradcell{0.178}{0.024}{#1}}
\newcommand{\cJSDbf}[1]{\gradcellbf{0.178}{0.024}{#1}}
\newcommand{\cRho}[1]{\gradcell{0.173}{0.912}{#1}}
\newcommand{\cRhobf}[1]{\gradcellbf{0.173}{0.912}{#1}}
\newcommand{\cAcc}[1]{\gradcell{0.620}{0.937}{#1}}
\newcommand{\cAccbf}[1]{\gradcellbf{0.620}{0.937}{#1}}
\newcommand{\cMAE}[1]{\gradcell{0.345}{0.066}{#1}}
\newcommand{\cMAEbf}[1]{\gradcellbf{0.345}{0.066}{#1}}

\newcommand\blankfootnote[1]{%
  \begingroup
  \renewcommand\thefootnote{}\footnote{#1}%
  \addtocounter{footnote}{-1}%
  \endgroup
}

\usepackage{fontspec}
\usepackage{xeCJK}
\newfontfamily\arabicfont{Amiri-Regular.ttf}[
    BoldFont = Amiri-Bold.ttf,
    ItalicFont = Amiri-Italic.ttf,
    BoldItalicFont = Amiri-BoldItalic.ttf,
    Scale=MatchLowercase,
    Script = Arabic 
]

\newcommand{\ar}[1]{\beginR{\arabicfont #1}\endR}

\newcommand{\rmv}[1]{}

\title{Behavioral and Representational Evidence of Binomial Ordering Preferences in Large Language Models}

\author{
\textbf{Zhiqing Yang}${}^{1,\dagger}$ \quad
\textbf{Yilun Liu}${}^{1,2,\dagger}$ \quad
\textbf{Yunpu Ma}${}^{1,2}$ \quad
\textbf{Volker Tresp}${}^{1,2}$ \quad
\textbf{Hinrich Sch\"utze}${}^{1,2}$\\
${}^{1}$\text{Ludwig Maximilian University of Munich} \quad
${}^{2}$\text{Munich Center for Machine Learning} \\
\normalsize \texttt{zhiqing.yang@campus.lmu.de} \quad \texttt{liuyilun2000@126.com} \quad
\normalsize ${}^{\dagger}$\textit{Equal contribution.}
}

\begin{document}
\maketitle
\input{latex/abstract}

\input{latex/introduction}

\input{latex/related}
\input{latex/methodology}
\input{latex/experiments}
\input{latex/conclusion}
\input{latex/limitations}
\bibliography{custom}

\FloatBarrier
\clearpage

\appendix
\label{sec:appendix}
\input{latex/appendixA}

\end{document}

%% file: latex/abstract.tex
\begin{abstract}

Large language models (LLMs) can readily reproduce conventional expressions, yet their ability to model gradient frequency distributions remains underexplored. 
We investigate this using linguistic binomials, such as \textit{men and women}, where both word permutations are grammatically valid but exhibit distinct, cross-linguistic variations in conventionality.
We formalize binomial ordering as a distributional alignment problem, and construct a multilingual dataset of 600 binomial pairs across 8 languages.
With categorical and distributional metrics, we measure and compare the corpus-derived preferences with model-induced ordering probabilities of 6 open-weight LLMs.
While models often behaviorally recover the dominant corpus-preferred order, particularly for strongly conventionalized pairs, they align less well with the exact corpus preference distributions. 
This suggests that apparent directional order overstates how faithfully LLMs capture the statistical nuances of language use. 
Sparse probing verifies that the concept of preference strength is partially encoded among middle-to-late layers, and steering along probe-derived directions alters model-induced ordering distributions, demonstrating that the statistical behavioral preference of LLMs can be mechanistically measured and manipulated via internal representations. 
\blankfootnote{\noindent
Code and data are publicly available at\\
\url{https://github.com/Zhi-qing-Yang/Linguistic-Binomials-in-Large-Language-Models}.}
\end{abstract}

%% file: latex/introduction.tex
\begin{figure*}[t]
    \centering
        \includegraphics[width=\linewidth]{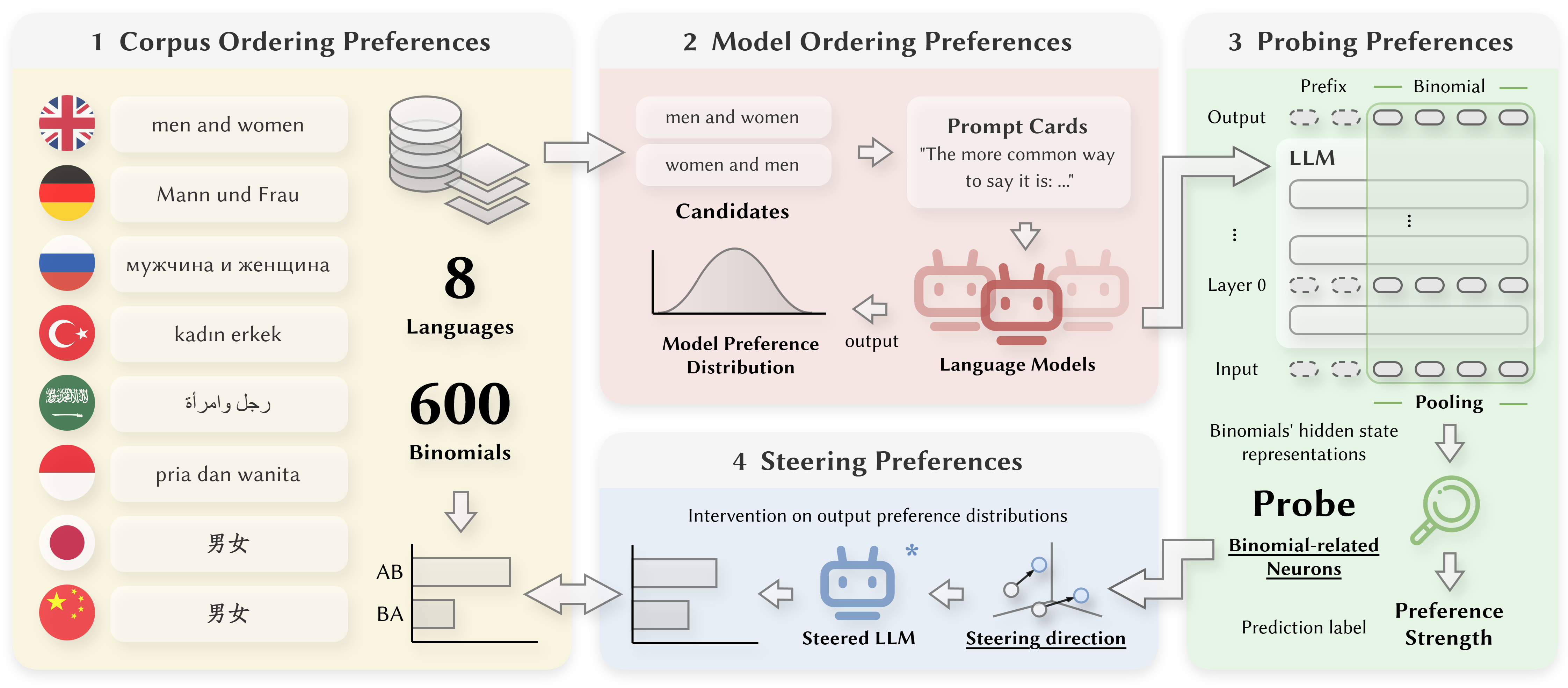}
        \vspace{-3ex}
        \caption{
    Pipeline for analyzing behavioral and representational binomial ordering preferences of LLMs. Corpus statistics define reference preferences, model scores induce predicted preferences, and representation-level probing and steering test whether these preferences are encoded and causally manipulable.}
        \vspace{-2ex}
        \label{fig:main}
\end{figure*}

\section{Introduction}

Large language models (LLMs) are increasingly examined for their ability to reflect human-like linguistic knowledge, including grammatical judgments, rare linguistic phenomena, and cross-lingual diversity~\citep{espinosa-anke-etal-2021-evaluating,madabushi2022semeval2022task2multilingual,de-luca-fornaciari-etal-2024-hard,Hu_2024,qiu-etal-2024-evaluating,misra2025languagemodelslearnrare,guo2025benchmarkinglinguisticdiversitylarge}. 
The majority of this work focuses on evaluating whether models distinguish acceptable from unacceptable forms, while natural language often involves gradient preferences among multiple alternatives that are all grammatically valid. 
This leaves open whether LLMs capture the distributional strength of conventional usage preferences beyond the binary direction.

Linguistic binomials provide an ideal test subject to isolate and examine this capability. 
Binomials are conventionalized two-word conjunctive phrases, whose components exhibit a preferred ordering that may vary across languages and specific lexical items~\citep{Malkiel1959StudiesII,mollin2012}. 
These preferences are rarely enforced by rigid rules, but are often shaped by a complex interplay of phonological, semantic, pragmatic, frequency-based, and sometimes socially mediated constraints~\citep{cooperross1975worldorder,benor2006,mollin2012,motschenbacher2013gentlemen}. 
Because binomials cleanly separate basic grammatical acceptability from the gradient strength of conventional ordering, they allow us to test whether LLMs faithfully mirror empirical usage patterns.
Prior work has shown that ordering preferences English LLMs behave are driven mainly by corpus statistical patterns~\citep{houghton-etal-2025-role}, yet it is still unexplored whether the model behavior distribution align with the corpus preferences, whether the alignment patterns generalize across languages and models, and whether the preferences are encoded in internal representations.

To address these gaps, we formalize binomial ordering as a distributional alignment problem, and construct a multilingual dataset comprising 600 binomial pairs across 8 languages, covering typologically diverse scripts and language families.
For each binomial pair, we estimate a corpus-derived ordering distribution from the relative frequencies of the two candidate orders, and compare this with a model-induced ordering distribution derived from the relative likelihoods assigned by LLMs. 
Alignment can then be assessed at two levels, whether the model prefers the right direction and whether it assigns the right strength.
Behavioral evaluation across 6 open-weight LLMs spanning different families and parameter scales reveals that while models recover the empirical preference direction reliably, especially for strongly conventionalized pairs,
the probability-level preference strength aligns less faithfully, with systematic variation across languages and model families.

To answer whether preference strength information has been absent from the model, or has been partially internalized but imperfectly expressed in output probabilities, we probe hidden states at each decoder layer to test whether the ordering preference strength is linearly encoded~\citep{alain2018understandingintermediatelayersusing,hewitt-manning-2019-structural}, 
and apply activation steering to test whether moving along the decoded direction shifts models' preferences~\citep{turner2024steeringlanguagemodelsactivation,panickssery2024steeringllama2contrastive}.
Results confirm that the gradient preference strength is partially recoverable at middle-to-late layers, and steering within internal representations has visible systematic behavioral consequences in model-induced ordering probabilities.

Our contributions are as follows:
\begin{enumerate}[itemsep=3pt, topsep=3pt, parsep=0pt, leftmargin=*, labelindent=0pt]
    \item We propose an evaluation protocol of binomial ordering preference for LLMs, with a dataset of 600 binomial pairs across 8 languages;
    \item We show that LLMs behaviorally align with the empirically preferred direction more reliably than the strength, and that strength can be representationally located and manipulated.
\end{enumerate}

%% file: latex/related.tex
\section{Related Work}
\label{sec:related}

Recent studies have evaluated whether LLMs capture linguistic knowledge across a variety of dimensions, including grammatical judgments~\citep{Hu_2024,qiu-etal-2024-evaluating,ide-etal-2025-make}, syntactic behavior~\citep{tenney2019bertrediscoversclassicalnlp}, rare constructions~\citep{misra2025languagemodelslearnrare}, and multilingual benchmarks~\citep{hu2020xtrememassivelymultilingualmultitask}.
These studies, however, rely on discrete contrasts or task-level accuracy, and less attention has been paid to gradient features where multiple alternatives are acceptable, but some are conventionally preferred to a measurable degree. 
For usage-based linguistic knowledge, the question is not only which form a model prefers, but whether preference strength also matches the empirical distribution. 

Binomial ordering has long been used to study how conventional preferences emerge from repeated language use. 
Earlier work demonstrates that speakers are sensitive to broad ordering tendencies~\citep{cooperross1975worldorder} as well as item-specific experience, making some pairs strongly associated with one order while others remain variable~\citep{mollin2012,morgan2015modeling}.
Similar to multi-word expressions, binomials involve familiar combinations rather than fully productive rules~\citep{pinker1979speakers,morgan2016abstract}. 
Human usage favors multiple plausible orderings to different degrees, making binomials a fundamentally gradient phenomenon, far more so than much Natural Language Processing subjects on conventionalized expressions. 
Recent work on English binomials suggests that LLM preferences are shaped by corpus-observed ordering patterns~\citep{houghton-etal-2025-role}. 
We extend this perspective by investigating how well model-induced preference strength behaviorally and representationally aligns with corpus-derived ordering distribution across languages.

Cross-lingual evidence suggests that preferences within a single language should not be assumed to transfer seamlessly to other languages. 
LLM performance varies with language resource levels, scripts, typological distance, and task formats~\citep{hu2020xtrememassivelymultilingualmultitask}, Furthermore, multilingual LLMs, even with superb benchmark performances, still exhibit distinct behaviors across languages and linguistic phenomena, ~\citep{lai2023chatgptenglishcomprehensiveevaluation,ahuja2023megamultilingualevaluationgenerative,guo2025benchmarkinglinguisticdiversitylarge}. For binomials, preferred orders may differ across languages, and languages closer in script or structure may align with model preferences differently from more distant ones. We therefore evaluate ordering preferences across eight typologically diverse languages.

Behavioral preferences derived from model output do not answer how well gradient preference is encoded in models' internal representations. 
Probing methods are commonly used to test whether linguistic properties, including syntactic structures~\citep{hewitt-manning-2019-structural,tenney2019bertrediscoversclassicalnlp}, semantics~\citep{alain2018understandingintermediatelayersusing}, and other intermediate features~\citep{gurnee2023finding, jiao-etal-2024-spin}, can be genuinely recovered from hidden states. 
Yet recoverability alone does not guarantee the encoded features to causally affect model behavior. 
Activation steering methods address this limitation by testing whether modifying internal representations systematically affects model outputs~\citep{turner2024steeringlanguagemodelsactivation,panickssery2024steeringllama2contrastive}. 
We combine these perspectives by examining whether binomial preference strength is linearly recoverable from hidden states and whether moving along the resulting direction changes the model-induced ordering distribution.

%% file: latex/methodology.tex
\section{Methodology}
\label{sec:method}

\subsection{Binomial Formalization}
\label{sec:formalization}

Binomials are conventionally ordered coordinate expressions whose ordering preferences are often probabilistically preferred rather than categorically grammatical \citep{morgan2015modeling}.
We represent each binomial as $(A,g,B)$, where $A$ and $B$ are conjuncts and $g$ is the
connective. 
Each binomial pair defines two surface continuations,
$y_{AB}=A\,g\,B$ and $y_{BA}=B\,g\,A$. 
Since these surfaces may be tokenized into different numbers of sub-word tokens, all model scores are computed over the tokenizer-specific token sequence sections $\mathbf{x}_{AB}$ and $\mathbf{x}_{BA}$.

\subsection{Corpus and Model Ordering Preferences}
\label{sec:corpus-dist}

For each binomial item $i$, let $n_{AB}^{(i)}$ and $n_{BA}^{(i)}$ be the corpus counts of the two orders.
Without loss of generality, we take  $A$--$B$ as the preferred order when $n_{AB}^{(i)} \geq n_{BA}^{(i)}$.
We estimate corpus ordering probability as the posterior mean of a Bernoulli rate under a Jeffreys prior \citep{jeffreys1946invariant}:
\begin{align*}    
p_{\mathrm{corpus}}^{(i)}
=
\frac{n_{AB}^{(i)}+0.5}
     {n_{AB}^{(i)}+n_{BA}^{(i)}+1}.
\end{align*}

\label{sec:model-dist}

For model preference, we derive from the likelihood that the language model assigns to each candidate ordering.
Each ordering is realized as a tokenizer-specific continuation $\mathbf{x}=(x_1,\ldots,x_T)$ following specific prefix context tokens $\mathbf{u}$.
We score a continuation by its autoregressive log-likelihood under the decoder-based language model $\mathcal{M}$~\citep{radford2018improving}:
\[
s(\mathbf{x}\mid\mathbf{u})
=
\sum_{t=1}^{T}
\log p_{\mathcal{M}}(x_t\mid \mathbf{u},x_{<t}),
\]
where $p_\mathcal{M}$ denotes the next-token distribution of $\mathcal{M}$.
For each item and prompt condition, we score the tokenized continuations $\mathbf{x}_{AB}$ and
$\mathbf{x}_{BA}$, obtaining $s_{AB}$ and $s_{BA}$. We then convert the score difference into a
model-induced ordering probability:
\[
p_{\mathrm{LLM}}^{(i)}=\sigma(s_{AB}-s_{BA}),\;
\sigma(z)=(1+e^{-z})^{-1}.
\]

\subsection{Prompt Conditions}
\label{sec:conditions}

Because prompt wording can influence model behavior, a single prompt may provide only a partial estimate of model preference \citep{jiang-etal-2020-know}.
Therefore, we score each candidate ordering under four language-specific prefix conditions. The \texttt{minimal} condition presents the binomial as a simple phrase continuation, without explicitly mentioning frequency or conventionality. The \texttt{frequency} condition frames the continuation as the more common way of saying the expression. The \texttt{discourse} condition frames the expression as something speakers habitually say, targeting usage in ordinary discourse. 
The \texttt{metalinguistic} condition directly asks for the conventional order of the pair.
We then pool scored instances across all four conditions before computing metrics, so that the evaluation is less dependent on any single elicitation context. 
Prefix templates used for all 8 languages are listed in Appendix~\ref{app:prompts}.

\subsection{Evaluation Metrics}
\label{sec:metrics}

For each language and model, we compare $p_{\mathrm{LLM}}$ with
$p_{\mathrm{corpus}}$ using Spearman correlation, pairwise accuracy, mean absolute error (MAE), and Jensen–Shannon divergence.
Spearman correlation measures whether the model and corpus rank binomial preferences
similarly \citep{spearman1904proof}:
\[
\rho
=
\mathrm{Spearman}\!\left(p_{\mathrm{LLM}},p_{\mathrm{corpus}}\right).
\]
Beyond this within-language model--corpus comparison, Spearman correlation is also utilized for cross-language corpus--corpus, model--corpus, and model--model comparisons. 
These correlations are computed over the same set of binomials, with prompt-conditioned model probabilities
averaged within each language before comparison.

Pairwise accuracy measures whether the model and corpus prefer the same order. Let
$\hat{o}_{\mathrm{LLM}}^{(i)}$ and $\hat{o}_{\mathrm{corpus}}^{(i)}$ denote the model-preferred
and corpus-preferred ordering, respectively. Then
\[
\mathrm{Acc}
=
\frac{1}{N}\sum_{i=1}^{N}
\mathbf{1}\!\left[
\hat{o}_{\mathrm{LLM}}^{(i)}=\hat{o}_{\mathrm{corpus}}^{(i)}
\right].
\]
Probability-level distance is measured with MAE:
\[
\mathrm{MAE}
=
\frac{1}{N}\sum_{i=1}^{N}
\left|p_{\mathrm{LLM}}^{(i)}-p_{\mathrm{corpus}}^{(i)}\right|.
\]
For Jensen–Shannon divergence \citep{lin1991divergence}, let
\[
\begin{gathered}
P^{(i)}=(p_{\mathrm{corpus}}^{(i)},\,1-p_{\mathrm{corpus}}^{(i)}),\\[3pt]
Q^{(i)}=(p_{\mathrm{LLM}}^{(i)},\,1-p_{\mathrm{LLM}}^{(i)}),\\[3pt]
M^{(i)}=\tfrac{1}{2}(P^{(i)}+Q^{(i)}).
\end{gathered}
\]
We compute
\begin{align*}
D_\mathrm{JS}(P^{(i)}\|Q^{(i)})
&=
\frac{1}{2}D_{\mathrm{KL}}(P^{(i)}\|M^{(i)})\\
&+
\frac{1}{2}D_{\mathrm{KL}}(Q^{(i)}\|M^{(i)}),
\end{align*}
with $D_\mathrm{KL}$ the Kullback–Leibler divergence
\begin{align*}
D_\mathrm{KL}(P\|Q)
=
\sum_{k}
P[k]\,\log\frac{P[k]}{Q[k]},
\end{align*}
where $P[k]$ and $Q[k]$  are the $k$-th components of the distributions.

\begin{table*}[t]
\centering
\small
\setlength{\tabcolsep}{8pt}
\begin{tabular}{l|cccccccc|c}
\toprule
& \multicolumn{9}{c}{\textbf{$\bm{D}_\mathbf{JS}$ $\downarrow$}} \\
\cmidrule(l{2pt}r{0pt}){2-10}
\textbf{Model} & \textbf{EN} & \textbf{DE} & \textbf{RU} & \textbf{ID} & \textbf{AR}
  & \textbf{TR} & \textbf{JA} & \textbf{ZH} & \textbf{Avg} \\
\midrule
Qwen3-4B     & \cJSD{.077} & \cJSD{.128} & \cJSD{.121} & \cJSD{.132} & \cJSD{.178} & \cJSD{.139} & \cJSD{.069} & \cJSDbf{.024} & \cJSD{.109} \\
Qwen3-14B    & \cJSD{.072} & \cJSD{.098} & \cJSD{.108} & \cJSD{.114} & \cJSD{.145} & \cJSD{.094} & \cJSDbf{.045} & \cJSD{.027} & \cJSD{.088} \\
Llama-3.2-3B & \cJSD{.062} & \cJSD{.103} & \cJSD{.124} & \cJSD{.113} & \cJSD{.167} & \cJSD{.110} & \cJSD{.082} & \cJSD{.048} & \cJSD{.102} \\
Llama-3.1-8B & \cJSDbf{.053} & \cJSDbf{.080} & \cJSDbf{.100} & \cJSDbf{.088} & \cJSDbf{.111}
             & \cJSDbf{.074} & \cJSD{.069} & \cJSD{.030} & \cJSDbf{.076} \\
Gemma-3-4B   & \cJSD{.100} & \cJSD{.149} & \cJSD{.134} & \cJSD{.133} & \cJSD{.178} & \cJSD{.118} & \cJSD{.070} & \cJSD{.045} & \cJSD{.116} \\
Gemma-3-12B  & \cJSD{.106} & \cJSD{.151} & \cJSD{.148} & \cJSD{.132} & \cJSD{.166} & \cJSD{.105} & \cJSD{.060} & \cJSD{.039} & \cJSD{.113} \\
\bottomrule
\end{tabular}
\caption{Mean Jensen--Shannon divergence $D_\mathrm{JS}$ between model-induced and corpus-derived ordering distributions across languages. Lower values indicate closer model--corpus agreement; Avg reports the mean $D_\mathrm{JS}$ across languages. Cells are shaded by value, with red indicating stronger alignment, and blue indicating weaker alignment.}
\label{tab:jsd}
\vspace{-2ex}
\end{table*}

\subsection{Representational Analysis and Intervention}
\label{sec:interp}

We complement behavioral evaluation with representation-level probing and intervention
experiments, testing whether corpus-derived preference strength is encoded in hidden states
\citep{alain2018understandingintermediatelayersusing,belinkov-etal-2017-neural,hewitt-manning-2019-structural}
and whether steering along probe-derived directions can alter the ordering preference of model output
\citep{ravfogel-etal-2020-null,turner2023activation,zou2025representationengineeringtopdownapproach}.

\paragraph{Sparse Probing.}
For each scored continuation, we extract decoder-layer representations $\mathbf{x}_{\ell,r}$ at layer $\ell$ with pooling over the continuation span $r\in\{\texttt{last},\texttt{mean}\}$ using either the final continuation token or the mean over continuation tokens. The probing target is
\[
y^{(i)}
=
\left|p_{\mathrm{corpus}}^{(i)}-0.5\right|.
\]
For each layer--representation pair $(\ell,r)$ we fit
\[
\hat{y}^{(i)}
=
\mathbf{w}_{\ell,r}^{\top}\tilde{\mathbf{x}}_{\ell,r}^{(i)}
+
b_{\ell,r},
\]
where $\tilde{\mathbf{x}}_{\ell,r}^{(i)}$ is representation  with z-score standardization applied in each dimension:
\[
\tilde{\mathbf{x}}_{\ell,r}^{(i)}
=
\frac{\mathbf{x}_{\ell,r}^{(i)}-\boldsymbol{\mu}_{\ell,r}}{\boldsymbol{\sigma}_{\ell,r}},
\]
with $\boldsymbol{\mu}_{\ell,r}$ and $\boldsymbol{\sigma}_{\ell,r}$ the per-dimension mean and standard deviation estimated on the training split.
Each layer's sparse probe
uses an $\ell_1$-regularized linear regression objective \citep{tibshirani1996regression},
\[
\min_{\mathbf{w},b}
\frac{1}{N}\sum_{i=1}^{N}
\left(
y^{(i)}-\mathbf{w}^{\top}\mathbf{x}_{\ell,r}^{(i)}-b
\right)^2
+
\alpha\|\mathbf{w}\|_1 .
\]
We carry out $K$-fold cross validation in our probing experiments. The detailed data and model configurations are specified in Section~\ref{sec:exp}.

\paragraph{Activation Steering.}
We reuse the $K$ sparse probes fit across the $K$ cross-validation folds.
For a selected layer--representation pair $(\ell,r)$, we derive a steering vector by
averaging the fold-specific probe coefficients and normalizing:
\[
\bar{\mathbf{w}}_{\ell,r}
=
\frac{1}{K}\sum_{k=1}^{K}\mathbf{w}_{\ell,r}^{(k)},
\qquad
\mathbf{v}_{\ell,r}
=
\frac{\bar{\mathbf{w}}_{\ell,r}}
     {\|\bar{\mathbf{w}}_{\ell,r}\|_2}.
\]
At inference time, we add this direction to the selected layer's hidden states:
\[
\mathbf{h}_{t,\ell}
\leftarrow
\mathbf{h}_{t,\ell}
+
\lambda\mathbf{v}_{\ell,r},
\]
where $t$ indexes sequence position and $\lambda$ is the intervention strength. 
This intervention follows the broader line of work using
linear directions in representation space to analyze or control model behavior
\citep{turner2023activation,zou2025representationengineeringtopdownapproach,marks2024geometrytruthemergentlinear}.
The two candidate continuations are then rescored to obtain the updated $p_{\mathrm{LLM}}^{(i)}$.
Selected layers, representation types, and intervention scales are reported in Section~\ref{sec:exp}. 

%% file: latex/experiments.tex
\begin{figure*}[t]
\centering
\includegraphics[width=0.7\textwidth]{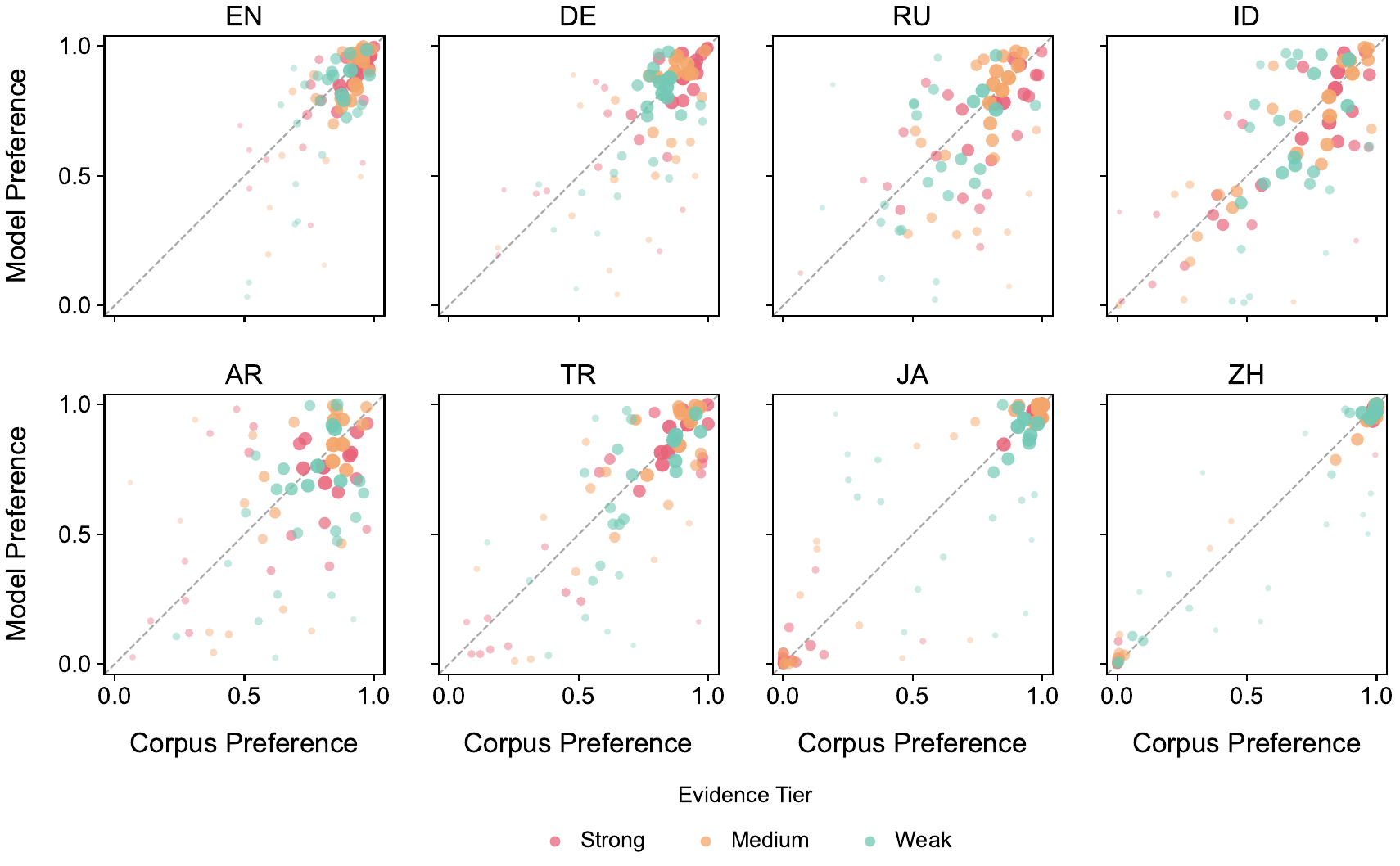}
\vspace{-1ex}
\caption{Item-level alignment between Llama-3.1-8B and corpus-derived ordering preferences
across languages. Each point corresponds to a language-specific binomial item, with
color indicating the corpus evidence tier based on the combined frequency of both
orders. The diagonal denotes perfect model--corpus agreement; larger deviations
indicate larger probability-level mismatches.}
\label{fig:behavioral-scatter}
\end{figure*}

\section{Experiments}
\label{sec:exp}

\subsection{Cross-linguistic Binomial Dataset}
\label{sec:dataset}

We construct a multilingual binomial dataset of 75 concept pairs shared across eight languages: English (EN), German (DE), Russian (RU), Indonesian (ID), Arabic (AR), Turkish (TR), Japanese (JA), and Chinese (ZH) . The pairs are selected to cover culturally common and cross-linguistically lexicalizable concepts, so that comparable expressions could be identified across all languages. 
We use conventional English binomials as the initial seed list, where the first conjunct corresponds to the commonly attested English order. This seed order is used only to define the shared concept inventory; corpus-based preferences are calculated independently for each target language.

For each language, we map the two concepts to language-specific lexical forms and define a set of frequent language-specific connectives for corpus querying. 
We then query the corresponding Sketch Engine\footnote{\url{http://www.sketchengine.eu}} \citep{kilgarriff2014sketch} TenTen corpora \citep{jakubicek2013tenten}, with the language-specific corpus sources listed in Appendix~\ref{app:corpora}, using CQL patterns \citep{jakubicek2010fast} that cover both $A$--$B$ and $B$--$A$ orders with these connectives. The resulting counts provide a language-specific reference distribution over the two possible orders.

Since raw corpus frequencies are not directly comparable across languages, we assign evidence tiers separately within each language with splitting binomial pairs into language-specific tertiles by the total number of occurrences of both the $A$--$B$ and $B$--$A$ orders, categorizing them as \texttt{weak}, \texttt{medium}, or \texttt{strong} evidence.

\subsection{Experiment Setup}
\label{sec:setup}
The evaluation dataset consists of multilingual binomial items paired with corpus-derived order preferences. Each item contains the scored order variants together with a corpus-based preference statistics $p_{\mathrm{corpus}}$ and an evidence tier reflecting the reliability of the corpus evidence. 
Our behavioral evaluation compares six open-weight multilingual causal LMs, Qwen3-4B/14B \citep{yang2025qwen3technicalreport}, Llama-3.2-3B \citep{meta2024llama32modelcard}, Llama-3.1-8B \citep{grattafiori2024llama3herdmodels}, and Gemma-3-4B/12B \citep{gemmateam2025gemma3technicalreport}, by measuring how closely $p_{\mathrm{LLM}}$ matches $p_{\mathrm{corpus}}$ across eight languages, four elicitation contexts (\texttt{minimal}, \texttt{frequency}, \texttt{discourse}, \texttt{metalinguistic}), and three evidence tiers.

For probing, we test where corpus-derived preference strength is linearly recoverable from Qwen3-4B hidden states, following Section~\ref{sec:interp}.
We use grouped 5-fold cross-validation with the binomial concept as the grouping unit, so that all language realizations of a given binomial fall in the same fold, and the probes are trained jointly over all languages to encourage recovering language-agnostic encoding of ordering-preference strength.
This prevents the model from being evaluated on the same binomial just in a different language than it was trained on, which would otherwise leak item-level information across folds.
Spearman correlation is used as the metric, and $\alpha \in \{0.01, 0.03, 0.05\}$ is varied to assess the alignment-sparsity tradeoff. Finally, we perform activation steering using the probe-derived direction from the last-token representation at layer 14. Intervention strengths are set to $\lambda \in \{0,\pm10,\pm20,\pm50\}$, with $\lambda=0$ denoting the original baseline, to test whether $p_{\mathrm{LLM}}$ shifts systematically. Appendix~\ref{app:graded-steering} additionally reports steering with the mean-pooled direction at layer 23.

\begin{figure}[t]
\centering
\includegraphics[width=\columnwidth]{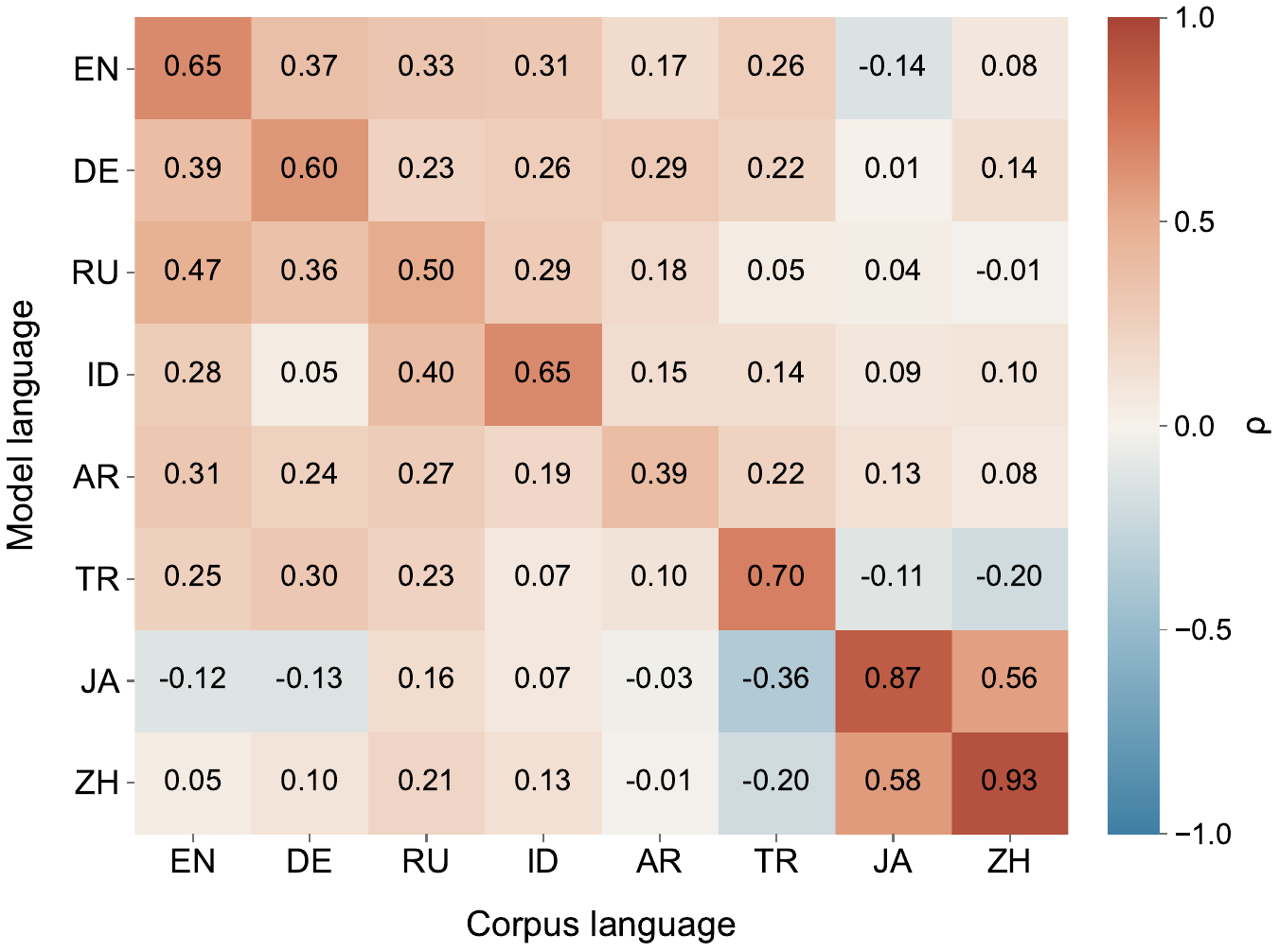}
\vspace{-3ex}
\caption{Model--corpus cross-linguistic correlation for Llama-3.1-8B.
Rows correspond to model-induced ordering distributions and columns to corpus-derived
ordering distributions.
Each cell reports the Spearman correlation $\rho$ between the model preferences and the corpus preferences.}
\label{fig:heatmap-model-corpus}
\end{figure}

\section{Results and Discussion}

\subsection{Behavioral Distributional Alignment}
\label{sec:behavioral-results}

\begin{figure*}[t]
\centering
\includegraphics[
  width=0.78\textwidth,
  height=0.32\textheight,
  keepaspectratio
]{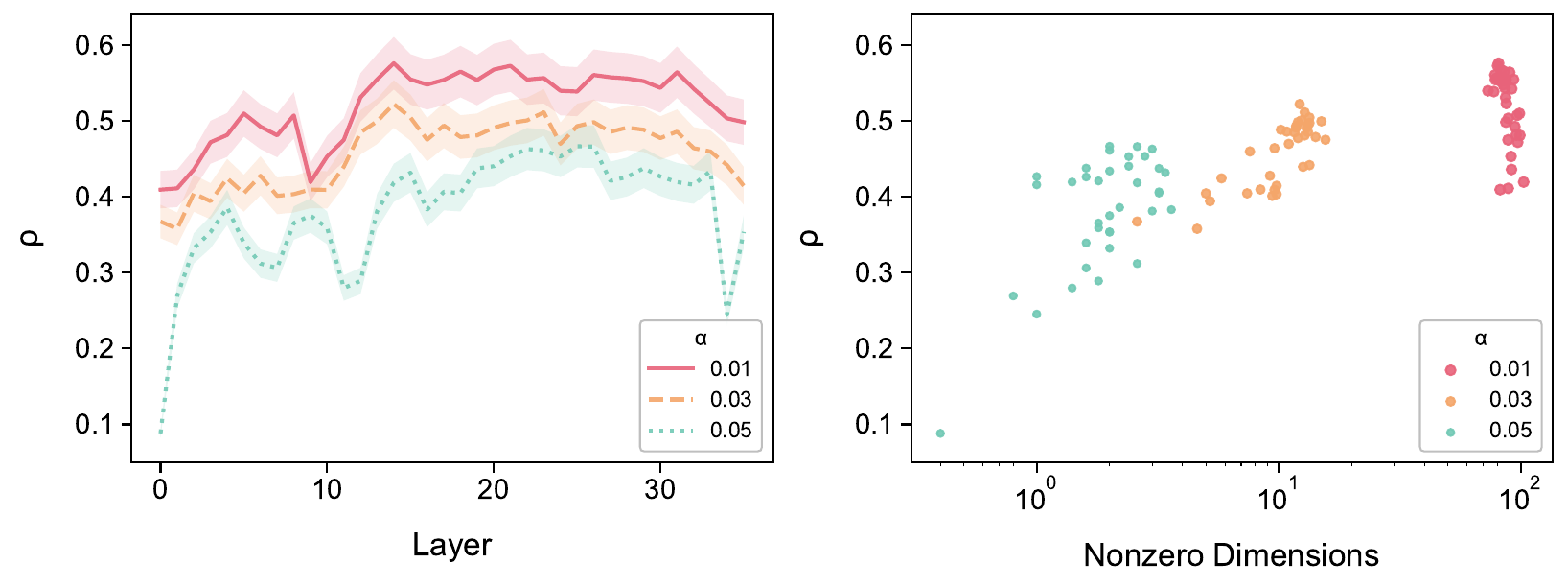}
\caption{
Sparse probing results for corpus-derived preference strength in Qwen3-4B using last-token binomial representations.
The y-axis shows Spearman correlation between predicted and corpus-derived preference strength; the left x-axis shows decoder layers, and the right x-axis shows the average number of nonzero Lasso coefficients.
Different Lasso regularization strengths are indicated with different colors, line styles (left), and marker sizes (right).
}
\label{fig:probe-lasso-last}
\end{figure*}

We first examine whether model-induced ordering distributions reproduce the empirical ordering distributions observed in corpora. 
Table 1 shows that Llama-3.1-8B achieves the lowest average divergence, and the larger models within each family  generally perform better. 
However, the alignment is only partial and the results vary significantly across languages.
ZH and JA show relatively better alignment across models, partly because many binomial items in these languages occupy the polarized ends of the corpus probability scale, as shown in Figure~\ref{fig:behavioral-scatter}. As corpus preferences are closer to categorical, the model can align well by simply recovering the dominant order. 
For EN, its corpus probabilities are less concentrated at the extremes, and its predictions remain comparatively close to the diagonal, indicating a match at the level of ordering probabilities and not only of preferred direction.
In other languages, particularly ID and TR, many points in Figure~\ref{fig:behavioral-scatter} fall in the correct preference quadrant as the corpus but remain far from the diagonal, indicating directionally consistent yet probability-displaced predictions across evidence tiers.
RU and AR show more diffuse patterns, indicating that neither direction nor strength is recovered cleanly. 
The Spearman correlation, pairwise accuracy, and MAE results in Appendix~\ref{app:behavioral-metrics} also further support that while models can often recover the corpus-preferred direction reliably, matching the strength of that preference remains consistently more difficult. 
This pattern is consistent with recent evidence that LLMs can capture broad linguistic tendencies without fully reproducing the distributional variation of
human-written text \citep{reinhart2025llms}.

Figure~\ref{fig:heatmap-model-corpus} further indicates that these mismatches are
structured across languages rather than random. 
Llama-3.1-8B often correlates most strongly with the matching target-language corpus along the diagnal in most rows, indicating that it preserves language-specific ordering information.
At the same time, the off-diagonal correlations are nonetheless substantial, showing that its induced distributions are not fully language-specific but appears to combine target-language evidence with broader cross-linguistic ordering tendencies. 
The additional corpus--corpus and model--model heatmaps in Appendix~\ref{app:behavioral-heatmaps} support this interpretation. 
Corpus preferences themselves exhibit cross-linguistic structure, and the model preserves
part of this structure while smoothing distinctions between language-specific
ordering distributions.

Together these results show that behavioral alignment is partial and uneven. Models recover the preferred direction of a binomial order reliably, especially for strongly conventionalized pairs, but do not consistently reproduce the gradient strength of that preference, and the gap varies systematically across languages. 
This dissociation between direction and strength motivates the representational analysis that follows.

\begin{figure*}[!t]
\centering
\includegraphics[width=0.9\textwidth]{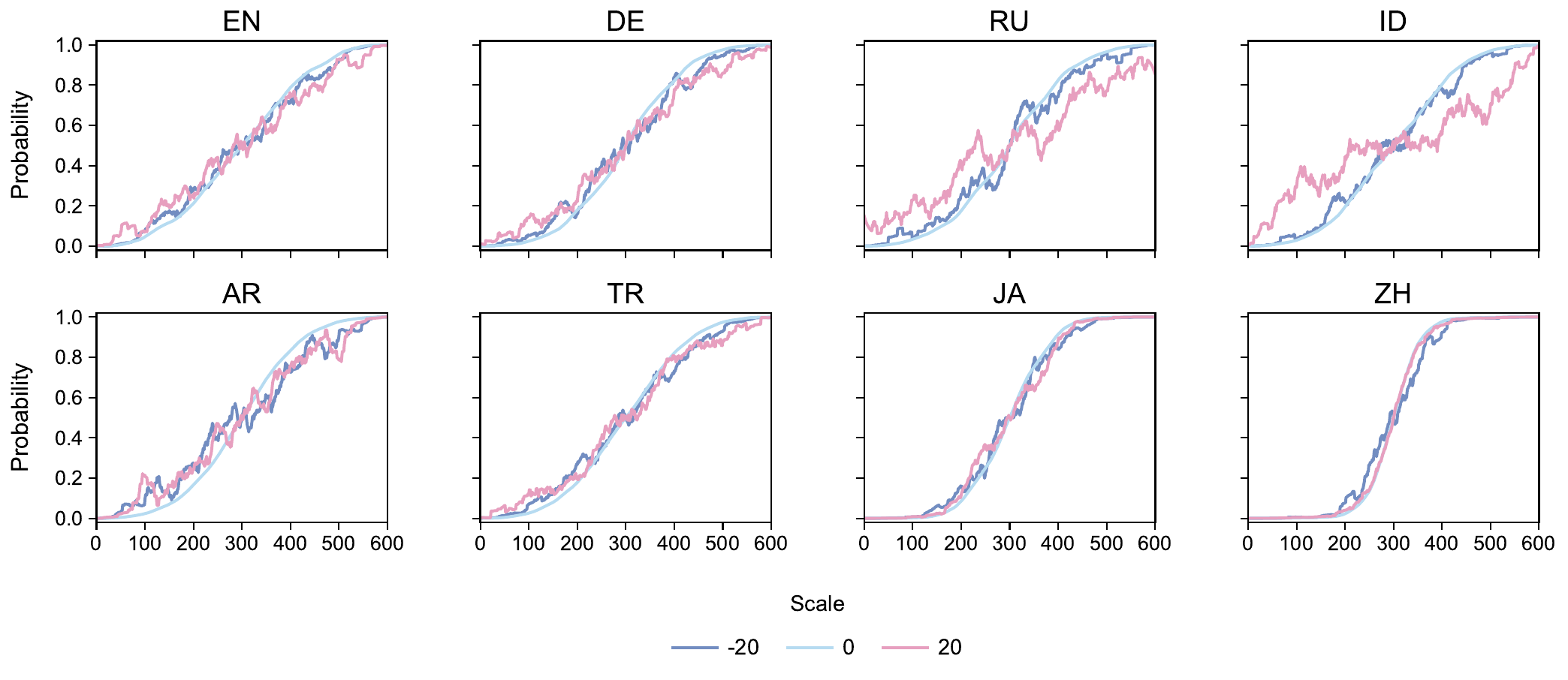}
\vspace{-1ex}
\caption{
Effects of steering on Qwen3-4B binomial order probabilities using the last-token steering vector at layer 14.
Each panel corresponds to one language; the x-axis indexes binomial orders sorted by their original probability, and the y-axis shows the model probability after steering.
Curves compare the original baseline ($\lambda=0$) with negative and positive steering ($\lambda=-20,+20$); additional scales are reported in Appendix~\ref{app:graded-steering}.
}
\label{fig:steering-curves}
\end{figure*}

\subsection{Probing Distribution}
\label{sec:probing-results}

We next test whether corpus-derived preference strength is linearly recoverable from
Qwen3-4B hidden states. Figure~\ref{fig:probe-lasso-last} shows a clear layer-wise
pattern for last-token binomial representations. 
Probing performance is weaker in the
beginning layers, rises through the middle of the model, and stays relatively high across a broad
middle-to-late region before declining near the final layers. 
This inverted-U profile, low at both ends and high in the middle, is the signature commonly reported in interpretability literature for various abstract and semantic properties, in contrast to surface features that peak early~\citep{tenney2019bertrediscoversclassicalnlp, liu2019linguistic, gurnee2023finding, tigges2023linear,  jiao-etal-2024-spin}. 
Its presence here is itself evidence that preference strength is an abstracted concept the model has actually acquired, rather than a shallow lexical correlate read off the input tokens.
The best probe reaches $\rho=0.576$ at layer 14 under weak
regularization, confirming a linearly recoverable signal for gradient preference strength and not only for the preferred ordering direction.

The signal is moreover compact with substantial sparsity.
At $\alpha=0.03$, the best last-token probe reaches $\rho=0.522$ at layer 14 with an average of only 12.2
nonzero dimensions across folds, a 0.48\% fraction of the 2,560-dimensional representation retaining most of the correlation in the weakly regularized probe. 
This indicates that preference strength is not diffusely smeared across the hidden state but concentrated in a low-dimensional subspace, which is what makes the following steering intervention in Section~\ref{sec:steering-results} feasible.
The mean-pooled representations in Figure~\ref{fig:probe-lasso-mean} show the same qualitative trend, with performance again peaking in the middle-to-late layers,  reaching $\rho=0.522$ at layer 23 under $\alpha=0.03$; layer-wise results for both representation types are reported in Table~\ref{tab:appendix-probing-layer-rho}. Thus, the effect is not an artifact of the last-token pooling choice.

\subsection{Steering Distribution}
\label{sec:steering-results}

Given that preference strength is decodable from a subspace, we finally investigate whether the probe-derived representational direction within models is behaviorally relevant. 
Can intervening along that subspace change the model's binomial ordering preferences?
Following prior work that uses representation-level interventions to test whether learned directions
are behaviorally meaningful \citep{ravfogel-etal-2020-null,turner2023activation,zou2025representationengineeringtopdownapproach},
we steer Qwen3-4B along the probe-derived direction at a range of scales and measure the resulting order-level probabilities. We use the last-token direction at layer 14, where the sparse probe achieves its best performance under $\alpha=0.03$; the mean-pooled direction at layer 23 is evaluated as a robustness check in Appendix~\ref{app:graded-steering}.

Figure~\ref{fig:steering-curves} confirms that the intervention changes the model's ordering preferences. 
For the original model, the rank-ordered probabilities form sigmoidal curves, with binomial ordering that the model significantly prefers and disprefers at the both end.
Steering changes the shape of these curves, especially at large positive scales. 
The effect is not a uniform shift, since steering does not push every order up or down. 
Instead, it flattens the curve, raising the probability of low-ranked orders and lowering that of high-ranked ones.

\begin{figure}[!t]
\centering
\includegraphics[width=\columnwidth]{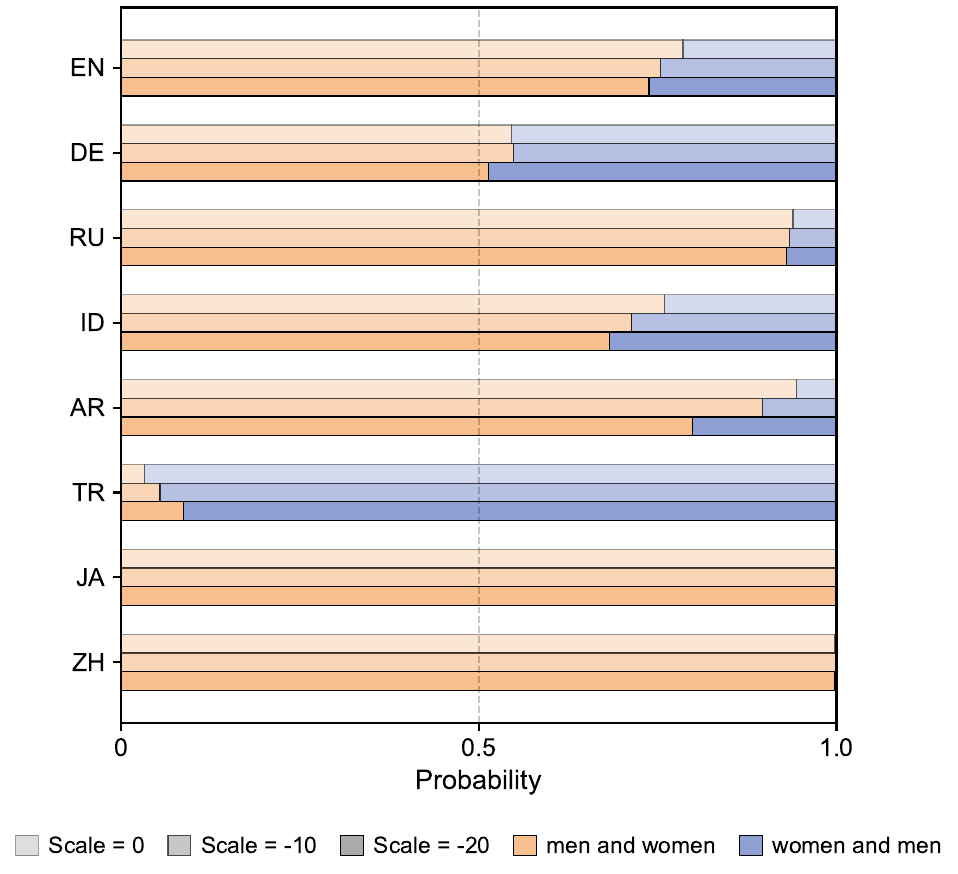}
\caption{
Steering case study for \textit{men and women} using the last-token steering vector at layer 14.
For each language, stacked bars compare the model probabilities of \textit{men and women} (orange) and \textit{women and men} (blue) at $\lambda=0,-10,-20$, with $\lambda=0$ denoting the original baseline.
}
\label{fig:steering-men-women}
\vspace{-2ex}
\end{figure}

The size of this effect differs across languages. 
ID and RU show the clearest
suppression effect under large positive steering, with the curve lifted on the left and lowered on the right as the model becomes less determined in its original ordering. 
preferences. 
EN follows the same tendency more mildly. 
In JA and ZH, the binomial ordering preferences are more extreme, with a ceiling effect where leaves less visible room for change, since a baseline already pinned near zero or one mechanically limits how far probabilities can move, even though steering still affects parts of the curve. 
This shows how the steering direction interacts with the baseline sharpness of each language's ordering preferences.

Figure~\ref{fig:steering-men-women} gives an illustrative view of the same phenomenon with
the binomial \textit{men and women}, making the cross language pattern easier to inspect in a
single item. 
Where the model initially assigns high probability to \textit{men and women},
steering lowers that probability. 
In TR, where the model initially prefers the reverse
order, steering weakens that reverse preference and correspondingly increases the probability of
\textit{men and women}. 
In JA and ZH the preference is pinned almost exactly at one, so steering produces little visible change.
The item-level pattern mirrors the curve-level result.
Steering mainly alters how strongly the model favors its currently preferred order in each language,  rather than imposing one fixed ordering preference direction order across languages.

Together, these results establish that the probe-derived representational directions has causal consequences in model behavior rather than mere artifacts of predictive correlations. 
This links the two halves of the representational analysis. The middle-to-late subspace from which preference strength is decodable is also the subspace through which that strength can be modulated, supporting the conclusion that gradient binomial preference is both internally encoded and causally active in model behavior.

%% file: latex/conclusion.tex
\section{Conclusion}

We confirm that LLMs capture binomial ordering preferences across languages, and we characterize the limits of this competence.
We formalize binomial ordering as a distributional alignment problem, comparing corpus-derived ordering preferences with model-induced distributions.
Behaviorally,  across languages and model families, LLMs reliably align with the corpus-preferred direction, yet capture the gradient strength of those preferences less consistently. 
Probing experiments show that corpus-derived preference strength is recoverable from hidden states, with the clearest signal appearing in middle-to-late layers, indicating that the preference strength is encoded in compact and linearly accessible forms. 
Steering further show that shifting hidden states along the decoded preference-strength direction alters model-induced ordering distributions, causally linking representational signals to model behavior. 
Our results reveal a subtle but informative gap in LLMs' linguistic knowledge, that models reliably recover which order is conventional, while how strongly that convention holds remains harder to capture.

%% file: latex/limitations.tex
\section*{Limitations}
\label{sec:limitations}

Corpus-derived ordering probabilities are a usage-based proxy for binomial
preferences, rather than direct human judgments. Although corpus frequencies reflect
conventional usage, they may also be shaped by genre, register, topic, and data
availability. Future work could compare model-induced preferences against human
acceptability or forced-choice judgments to better separate corpus-level conventions
from speaker-level preferences.

Estimating these probabilities from web corpora introduces additional sources of
variation. Although we use language-specific lexical forms and connectives, the
resulting counts may be affected by tokenization, morphology, orthographic variation,
and uneven domain coverage. Such issues are especially relevant in cross-linguistic
comparison, where the same binomial concept may not be equally lexicalized or equally
frequent across languages. For this reason, our results should be interpreted cautiously as
alignment with estimated corpus distributions, rather than an exhaustive record of
binomial usage.

Dataset construction also shapes the scope of the analysis. We use 75 concept pairs
shared across eight languages, with the concept inventory initially seeded from
conventional English binomials. This design makes cross-linguistic comparison
tractable, but may underrepresent binomial preferences specific to individual languages
or cultural contexts. The preferred order is nevertheless estimated independently for
each target language, so the evaluation does not assume that English ordering
preferences transfer to other languages.

At the representation level, our analyses are narrower in scope than the behavioral
evaluation. Probing and steering provide evidence that corpus-derived preference
strength is reflected in hidden states and related to model-induced ordering
probabilities, but the generality of this representation--behavior link across
architectures, training regimes, and intervention settings remains open. Extending
these analyses to more models and more exhaustive interventions would possibly better clarify how stable this link is across languages and model families.

\section*{Ethics Considerations}

This work studies linguistic binomial ordering preferences in large language models and does not target sensitive personal data or individual users.
Nevertheless, several ethical considerations merit attention.

The web corpora used to derive ordering preferences may reflect demographic, cultural, and socioeconomic biases present in large-scale online text.
Ordering preferences extracted from such corpora are therefore proxies for the usage patterns of populations overrepresented on the web, and may not reflect the preferences of speakers from underrepresented communities or linguistic varieties.
Conclusions drawn from corpus-derived preferences should accordingly be interpreted as reflecting dominant usage conventions rather than universal speaker knowledge.

The open-weight LLMs evaluated in this work were trained on large internet corpora and may themselves encode or amplify societal biases.
Binomial ordering is not inherently harmful, but some pairs in our dataset involve socially salient terms, such as gendered or culturally marked expressions, whose conventional ordering may reflect historical power asymmetries.
We do not intervene on or reinforce such orderings beyond measuring model behavior, but downstream applications that rely on model-preferred orderings in generated text should be aware of this potential.

The probing and steering experiments demonstrate that model-induced ordering preferences can be manipulated through representational intervention.
While we apply this methodology to study gradient preference strength, the same techniques could in principle be used to systematically bias model outputs toward non-conventional or socially skewed orderings.
We therefore caution against deploying probe-derived steering vectors in production systems without careful evaluation of downstream effects, particularly for pairs involving socially sensitive concepts.

We do not conduct dedicated bias audits or fairness evaluations on the models or corpora used.
Downstream users who adapt our methodology or dataset for applied purposes are encouraged to perform appropriate task-specific evaluations before deployment.

\section*{Acknowledgments}
The authors gratefully acknowledge the support and resources provided at the Center for Information and Language Processing (CIS), Ludwig Maximilian University of Munich, by the German Research Foundation (DFG, grant SCHU 2246/141). This work is also supported by the Munich Center for Machine Learning (MCML).

%% file: latex/appendixA.tex
\appendix

\section{Data Source and Prompt Details}
\label{app:data-prompts}

\subsection{Corpora}
\label{app:corpora}

Table~\ref{tab:corpora} lists the Sketch Engine TenTen corpora used to extract
order counts and estimate corpus-derived ordering preferences for each language.

\begin{table}[h]
\centering
\small
\setlength{\tabcolsep}{4pt}
\begin{tabularx}{\columnwidth}{@{}>{\centering\arraybackslash}X>{\centering\arraybackslash}X@{}}
\toprule
\textbf{Language} & \textbf{Corpus} \\
\midrule
EN & enTenTen21 \\
DE & deTenTen23 \\
RU & ruTenTen20 \\
ID & idTenTen24 \\
AR & arTenTen24 \\
TR & trTenTen20 \\
JA & jpTenTen11 \\
ZH & zhTenTen17 (Simplified) \\
\bottomrule
\end{tabularx}
\caption{Language-specific TenTen corpora used to estimate corpus-derived ordering preferences.}
\label{tab:corpora}
\end{table}

\subsection{Prefix Templates}
\label{app:prompts}

Table~\ref{tab:prompt-templates} lists the language-specific prefixes used for the
four elicitation contexts. During scoring, each candidate ordering is appended to the
corresponding prefix.

\begin{table*}[t]
\centering
\small
\setlength{\tabcolsep}{4pt}
\renewcommand{\arraystretch}{1.3}
\begin{tabular}{>{\centering\arraybackslash}m{0.035\textwidth}|
>{\raggedright\arraybackslash}m{0.15\textwidth}
>{\raggedright\arraybackslash}m{0.24\textwidth}
>{\raggedright\arraybackslash}m{0.12\textwidth}
>{\raggedright\arraybackslash}m{0.37\textwidth}
}
\toprule
& \textbf{minimal}
& \textbf{frequency}
& \textbf{discourse} 
& \textbf{metalinguistic} \\
\midrule
EN &
The phrase goes: &
The more common way to say it is: &
People always say: &
In this language, the conventional order for this pair is: \\[9pt]

DE &
Die Redewendung lautet: &
Die häufigere Art, es zu sagen, lautet: &
Man sagt immer: &
In dieser Sprache lautet die übliche Reihenfolge dieses Paares: \\[9pt]

RU &
Выражение звучит так: &
Более распространённый способ сказать это: &
Всегда говорят: &
В этом языке привычный порядок для этой пары таков: \\[9pt]

ID &
Ungkapan ini adalah: &
Cara yang lebih umum untuk mengatakannya adalah: &
Orang selalu bilang: &
Dalam bahasa ini, urutan lazim untuk pasangan ini adalah: \\[9pt]

AR &
\raggedleft \ar{هذه العبارة هي:} &
\raggedleft \ar{الطريقة الأكثر شيوعًا للتعبير عن ذلك هي:} &
\raggedleft \ar{يقولون دائمًا:} &
\raggedleft\arraybackslash \ar{في هذه اللغة، الترتيب المعتاد لهذا الزوج هو:} \\[9pt]

TR &
Bu ifade şöyledir: &
Daha yaygın söyleniş şekli şöyledir: &
Hep derler ki: &
Bu dilde bu çiftin alışılmış sırası şöyledir: \\[9pt]

JA &
この表現は：&
より一般的な言い方は：&
よく言う：&
この言語でこの語の組の慣用的な順序は：\\[9pt]

ZH &
这个说法是：&
更常见的说法是：&
大家都说：&
在这种语言中，这对词的惯用顺序是：\\
\bottomrule
\end{tabular}
\caption{Language-specific prefixes used for the four elicitation contexts.}
\label{tab:prompt-templates}
\end{table*}

\section{Behavioral Analyses}
\label{app:behavioral}

\subsection{Evaluation Metrics}
\label{app:behavioral-metrics}

In addition to the Jensen--Shannon divergence results, Tables~\ref{tab:appendix-rho}--\ref{tab:appendix-mae} report Spearman correlation for rank-level agreement, pairwise accuracy for preferred-order direction, and MAE for probability-level deviation.

\begin{table*}[t]
\centering
\small
\setlength{\tabcolsep}{8pt}
\begin{tabular}{l|cccccccc|c}
\toprule
& \multicolumn{9}{c}{\textbf{Spearman $\rho$ $\uparrow$}} \\
\cmidrule(l{2pt}r{0pt}){2-10}
\textbf{Model} & \textbf{EN} & \textbf{DE} & \textbf{RU} & \textbf{ID} & \textbf{AR}
  & \textbf{TR} & \textbf{JA} & \textbf{ZH} & \textbf{Avg} \\
\midrule
Qwen3-4B     & \cRho{.513} & \cRho{.396} & \cRho{.286} & \cRho{.373} & \cRho{.173} & \cRho{.341} & \cRho{.850} & \cRho{.903} & \cRho{.479} \\
Qwen3-14B    & \cRho{.438} & \cRho{.434} & \cRho{.375} & \cRho{.455} & \cRho{.311} & \cRho{.543} & \cRhobf{.899} & \cRhobf{.912} & \cRho{.546} \\
Llama-3.2-3B & \cRho{.520} & \cRho{.322} & \cRho{.341} & \cRho{.439} & \cRho{.240} & \cRho{.522} & \cRho{.783} & \cRho{.873} & \cRho{.505} \\
Llama-3.1-8B & \cRhobf{.566} & \cRhobf{.504} & \cRho{.411} & \cRhobf{.565} & \cRhobf{.331}
             & \cRhobf{.634} & \cRho{.843} & \cRho{.905} & \cRhobf{.595} \\
Gemma-3-4B   & \cRho{.426} & \cRho{.313} & \cRho{.245} & \cRho{.392} & \cRho{.193} & \cRho{.526} & \cRho{.867} & \cRho{.857} & \cRho{.477} \\
Gemma-3-12B  & \cRho{.511} & \cRho{.372} & \cRhobf{.439} & \cRho{.431} & \cRho{.292} & \cRho{.563} & \cRho{.857} & \cRho{.861} & \cRho{.541} \\
\bottomrule
\end{tabular}
\caption{Spearman correlation $\rho$ between model-induced and corpus-derived ordering
preferences across languages. Higher values indicate closer rank-level model--corpus
agreement.}
\label{tab:appendix-rho}
\end{table*}

\begin{table*}[t]
\centering
\small
\setlength{\tabcolsep}{8pt}
\begin{tabular}{l|cccccccc|c}
\toprule
& \multicolumn{9}{c}{\textbf{Pairwise accuracy $\uparrow$}} \\
\cmidrule(l{2pt}r{0pt}){2-10}
\textbf{Model} & \textbf{EN} & \textbf{DE} & \textbf{RU} & \textbf{ID} & \textbf{AR}
  & \textbf{TR} & \textbf{JA} & \textbf{ZH} & \textbf{Avg} \\
\midrule
Qwen3-4B     & \cAcc{.783} & \cAcc{.707} & \cAccbf{.703} & \cAcc{.750} & \cAcc{.620} & \cAcc{.710} & \cAcc{.833} & \cAccbf{.937} & \cAcc{.755} \\
Qwen3-14B    & \cAccbf{.817} & \cAccbf{.770} & \cAcc{.673} & \cAcc{.777} & \cAcc{.650} & \cAcc{.737} & \cAccbf{.900} & \cAcc{.927} & \cAcc{.781} \\
Llama-3.2-3B & \cAcc{.810} & \cAcc{.683} & \cAcc{.630} & \cAcc{.730} & \cAcc{.633} & \cAcc{.763} & \cAcc{.820} & \cAcc{.910} & \cAcc{.748} \\
Llama-3.1-8B & \cAcc{.810} & \cAccbf{.770} & \cAcc{.697} & \cAccbf{.787} & \cAccbf{.687}
             & \cAccbf{.807} & \cAcc{.817} & \cAcc{.920} & \cAccbf{.787} \\
Gemma-3-4B   & \cAcc{.770} & \cAcc{.677} & \cAcc{.637} & \cAcc{.700} & \cAcc{.657} & \cAcc{.697} & \cAcc{.830} & \cAcc{.890} & \cAcc{.732} \\
Gemma-3-12B  & \cAcc{.723} & \cAcc{.657} & \cAcc{.643} & \cAcc{.710} & \cAcc{.620} & \cAcc{.770} & \cAcc{.837} & \cAcc{.887} & \cAcc{.731} \\
\bottomrule
\end{tabular}
\caption{Pairwise accuracy between model-induced and corpus-derived ordering
preferences across languages. Higher values indicate that the model more often assigns
higher probability to the corpus-preferred order.}
\label{tab:appendix-acc}
\end{table*}

\begin{table*}[t]
\centering
\small
\setlength{\tabcolsep}{8pt}
\begin{tabular}{l|cccccccc|c}
\toprule
& \multicolumn{9}{c}{\textbf{MAE $\downarrow$}} \\
\cmidrule(l{2pt}r{0pt}){2-10}
\textbf{Model} & \textbf{EN} & \textbf{DE} & \textbf{RU} & \textbf{ID} & \textbf{AR}
  & \textbf{TR} & \textbf{JA} & \textbf{ZH} & \textbf{Avg} \\
\midrule
Qwen3-4B     & \cMAE{.191} & \cMAE{.270} & \cMAE{.269} & \cMAE{.282} & \cMAE{.345} & \cMAE{.282} & \cMAE{.155} & \cMAEbf{.066} & \cMAE{.232} \\
Qwen3-14B    & \cMAE{.184} & \cMAE{.225} & \cMAE{.255} & \cMAE{.254} & \cMAE{.298} & \cMAE{.230} & \cMAEbf{.115} & \cMAE{.070} & \cMAE{.204} \\
Llama-3.2-3B & \cMAE{.172} & \cMAE{.247} & \cMAE{.291} & \cMAE{.263} & \cMAE{.320} & \cMAE{.244} & \cMAE{.175} & \cMAE{.105} & \cMAE{.227} \\
Llama-3.1-8B & \cMAEbf{.156} & \cMAEbf{.208} & \cMAEbf{.247} & \cMAEbf{.226} & \cMAEbf{.266}
             & \cMAEbf{.200} & \cMAE{.161} & \cMAE{.078} & \cMAEbf{.193} \\
Gemma-3-4B   & \cMAE{.217} & \cMAE{.291} & \cMAE{.294} & \cMAE{.282} & \cMAE{.330} & \cMAE{.262} & \cMAE{.153} & \cMAE{.097} & \cMAE{.241} \\
Gemma-3-12B  & \cMAE{.233} & \cMAE{.290} & \cMAE{.311} & \cMAE{.277} & \cMAE{.321} & \cMAE{.238} & \cMAE{.134} & \cMAE{.087} & \cMAE{.236} \\
\bottomrule
\end{tabular}
\caption{Mean absolute error (MAE) between model-induced and corpus-derived ordering
distributions across languages. Lower values indicate closer probability-level
model--corpus agreement.}
\label{tab:appendix-mae}
\end{table*}

\subsection{Cross-linguistic Preference Correlations}
\label{app:behavioral-heatmaps}

We use cross-linguistic Spearman correlation heatmaps to compare ordering-preference vectors across languages over the shared binomial set. Corpus--corpus correlations serve as an empirical reference for cross-language similarity. Model--corpus correlations test whether each model language aligns most strongly with its corresponding corpus language, while model--model correlations characterize the model's internal cross-language preference similarity.

\begin{figure*}[t]
\centering
\begin{subfigure}[t]{0.48\textwidth}
    \centering
    \includegraphics[width=\linewidth]{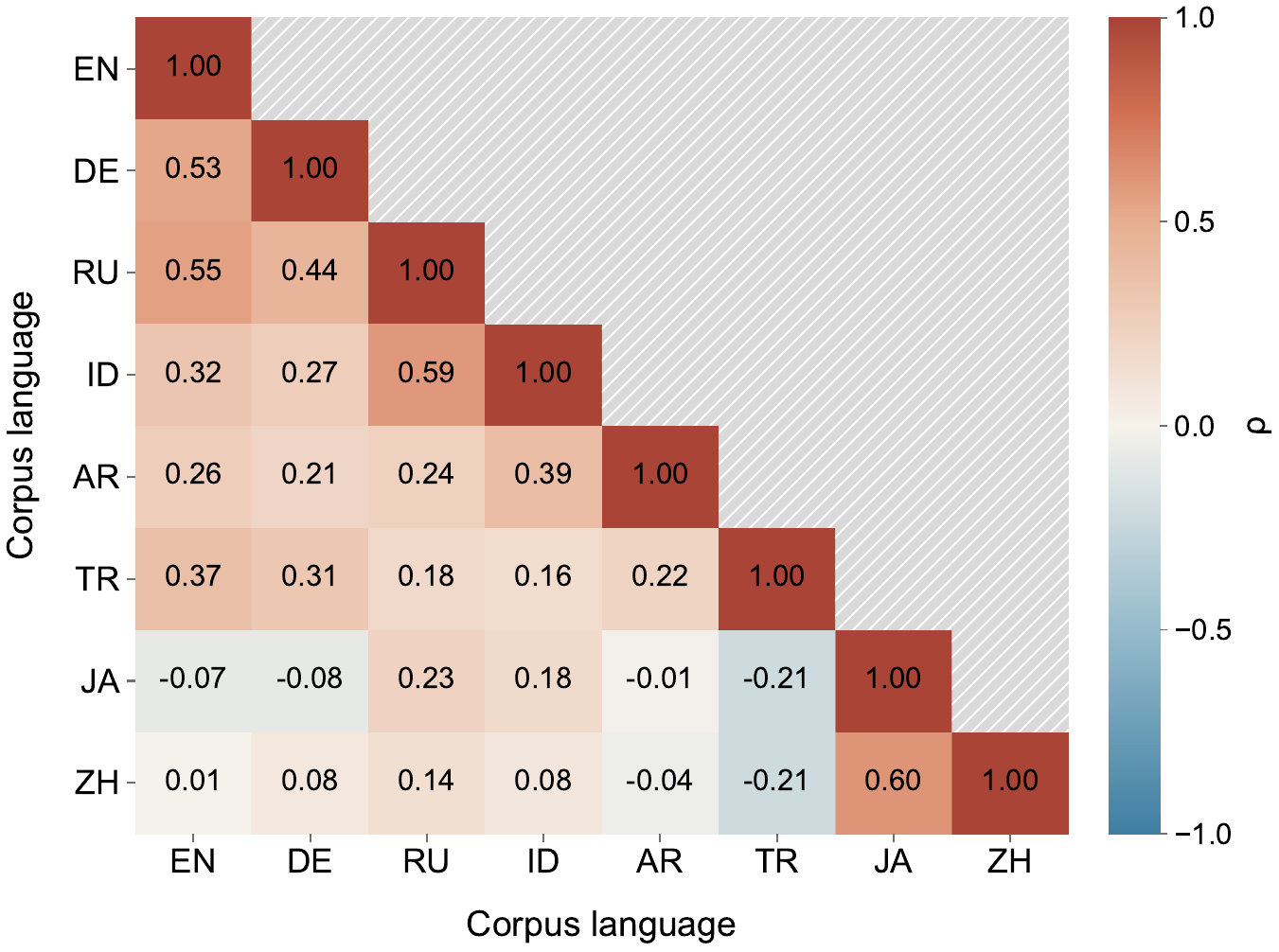}
    \caption{$p_{\text{corpus}}$ vs.\ $p_{\text{corpus}}$}
    \label{fig:appendix-heatmap-corpus-corpus}
\end{subfigure}
\hfill
\begin{subfigure}[t]{0.48\textwidth}
    \centering
    \includegraphics[width=\linewidth]{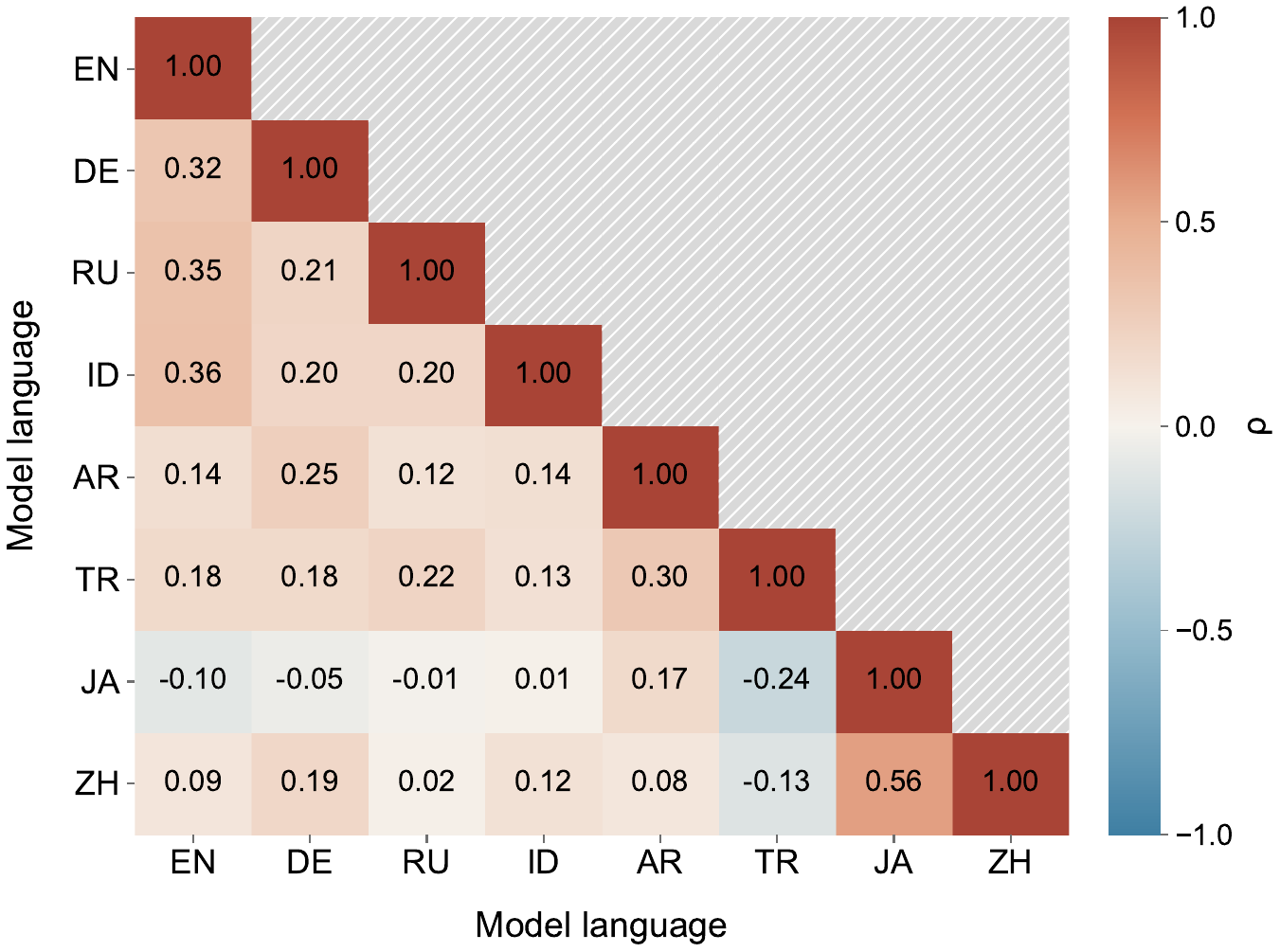}
    \caption{$p_{\text{LLM}}$ vs.\ $p_{\text{LLM}}$}
    \label{fig:appendix-heatmap-llama8b-model-model}
\end{subfigure}
\caption{Corpus reference and Llama-3.1-8B cross-linguistic structure.
Each cell reports the Spearman correlation between ordering-preference vectors
computed over shared binomial items. Left: corpus--corpus correlations, which
serve as the empirical reference for cross-linguistic similarity. Right:
model--model correlations for Llama-3.1-8B, showing the similarity structure
among the model's language-specific ordering preferences.}
\label{fig:appendix-crosslingual-heatmaps}
\end{figure*}

\begin{figure*}[t]
\centering
\begin{subfigure}[t]{0.48\textwidth}
    \centering
    \includegraphics[width=\linewidth]{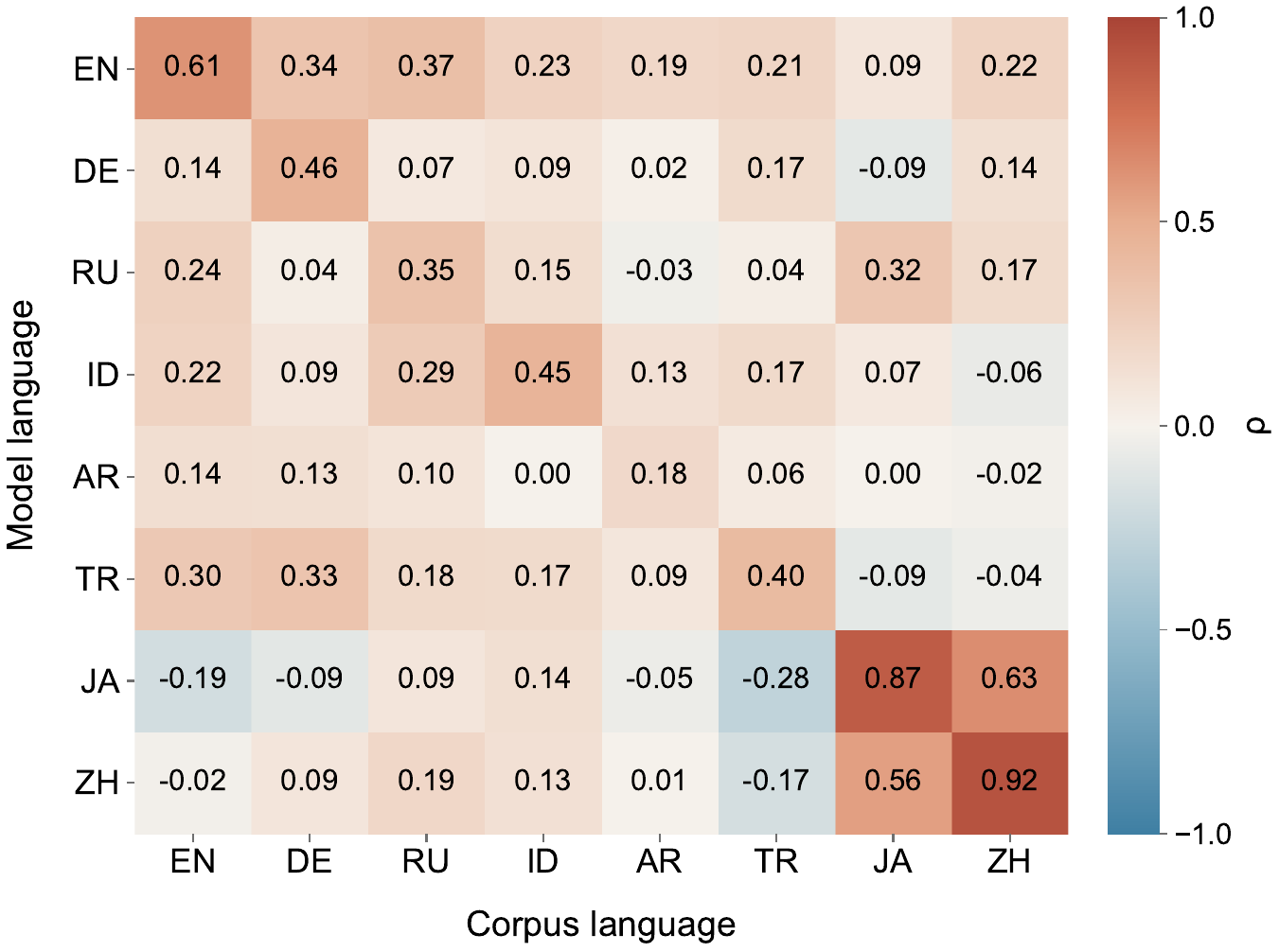}
    \caption{$p_{\text{LLM}}$ vs.\ $p_{\text{corpus}}$}
\end{subfigure}
\hfill
\begin{subfigure}[t]{0.48\textwidth}
    \centering
    \includegraphics[width=\linewidth]{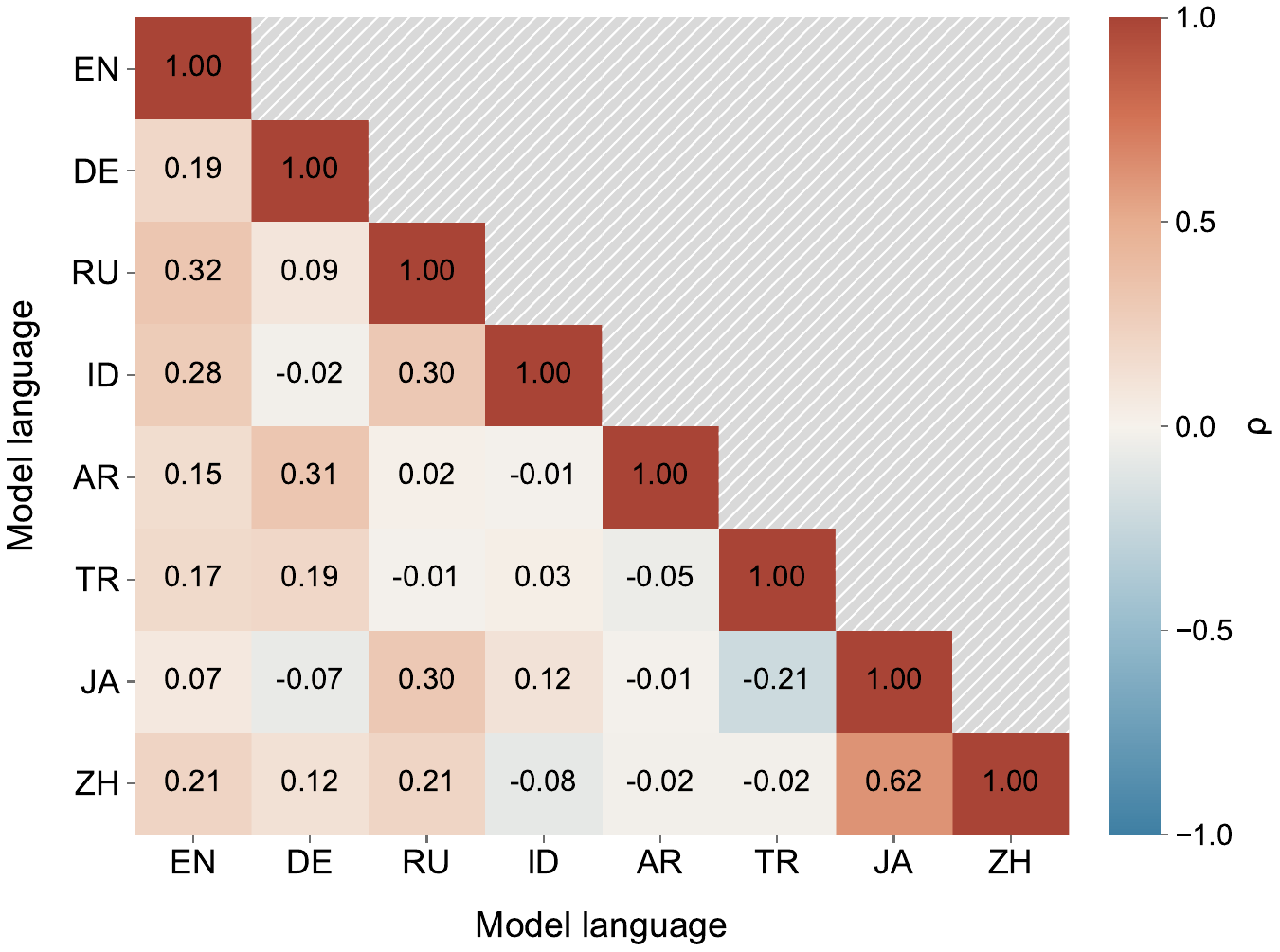}
    \caption{$p_{\text{LLM}}$ vs.\ $p_{\text{LLM}}$}
\end{subfigure}
\caption{Cross-linguistic alignment diagnostics for Qwen3-4B.
Left: model--corpus correlations. Right: model--model correlations.
Each cell reports Spearman correlation over shared binomial items.}
\label{fig:appendix-heatmaps-qwen4b}
\end{figure*}

\begin{figure*}[t]
\centering
\begin{subfigure}[t]{0.48\textwidth}
    \centering
    \includegraphics[width=\linewidth]{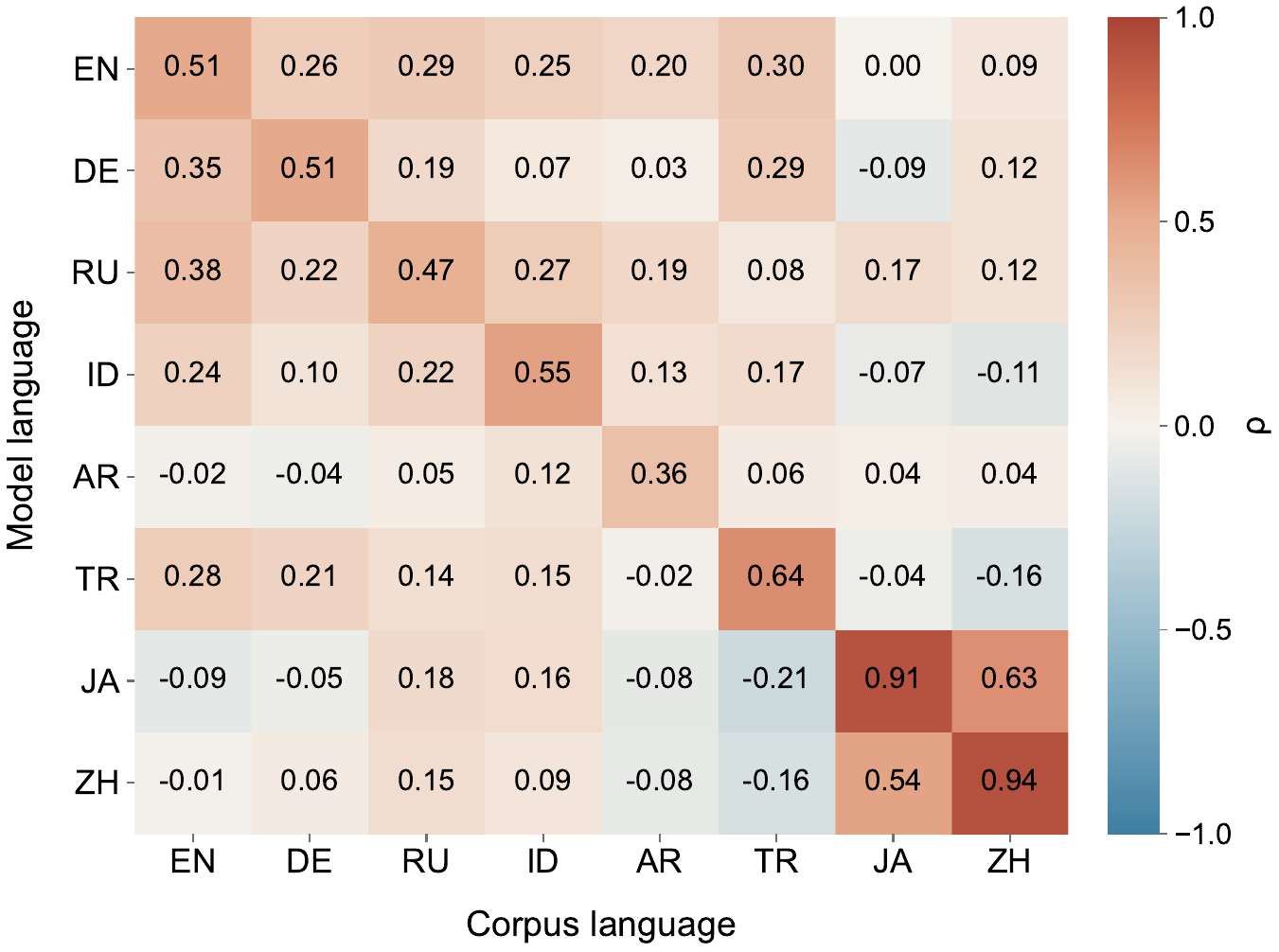}
    \caption{$p_{\text{LLM}}$ vs.\ $p_{\text{corpus}}$}
\end{subfigure}
\hfill
\begin{subfigure}[t]{0.48\textwidth}
    \centering
    \includegraphics[width=\linewidth]{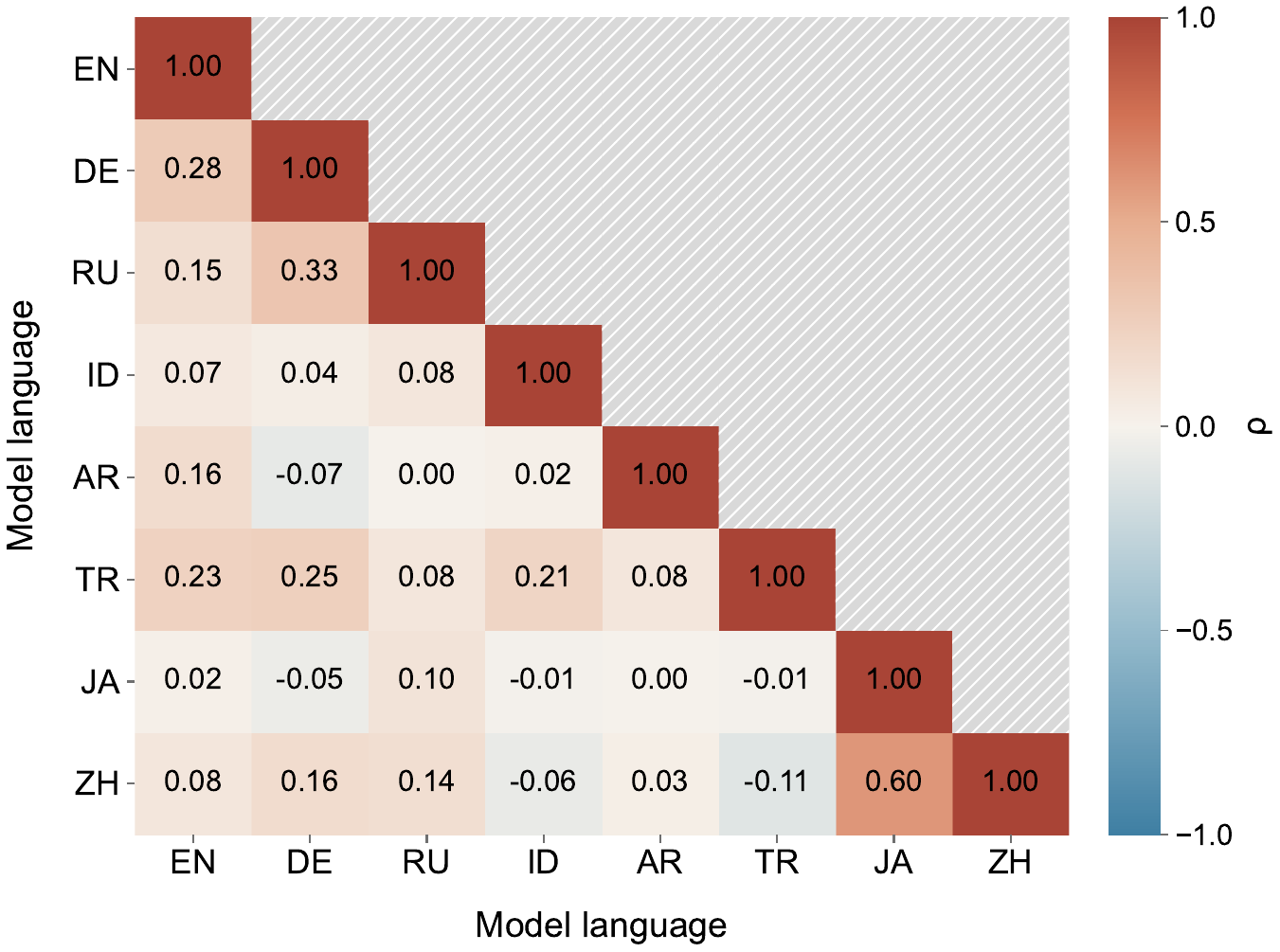}
    \caption{$p_{\text{LLM}}$ vs.\ $p_{\text{LLM}}$}
\end{subfigure}
\caption{Cross-linguistic alignment diagnostics for Qwen3-14B.
Left: model--corpus correlations. Right: model--model correlations.
Each cell reports Spearman correlation over shared binomial items.}
\label{fig:appendix-heatmaps-qwen14b}
\end{figure*}

\begin{figure*}[t]
\centering
\begin{subfigure}[t]{0.48\textwidth}
    \centering
    \includegraphics[width=\linewidth]{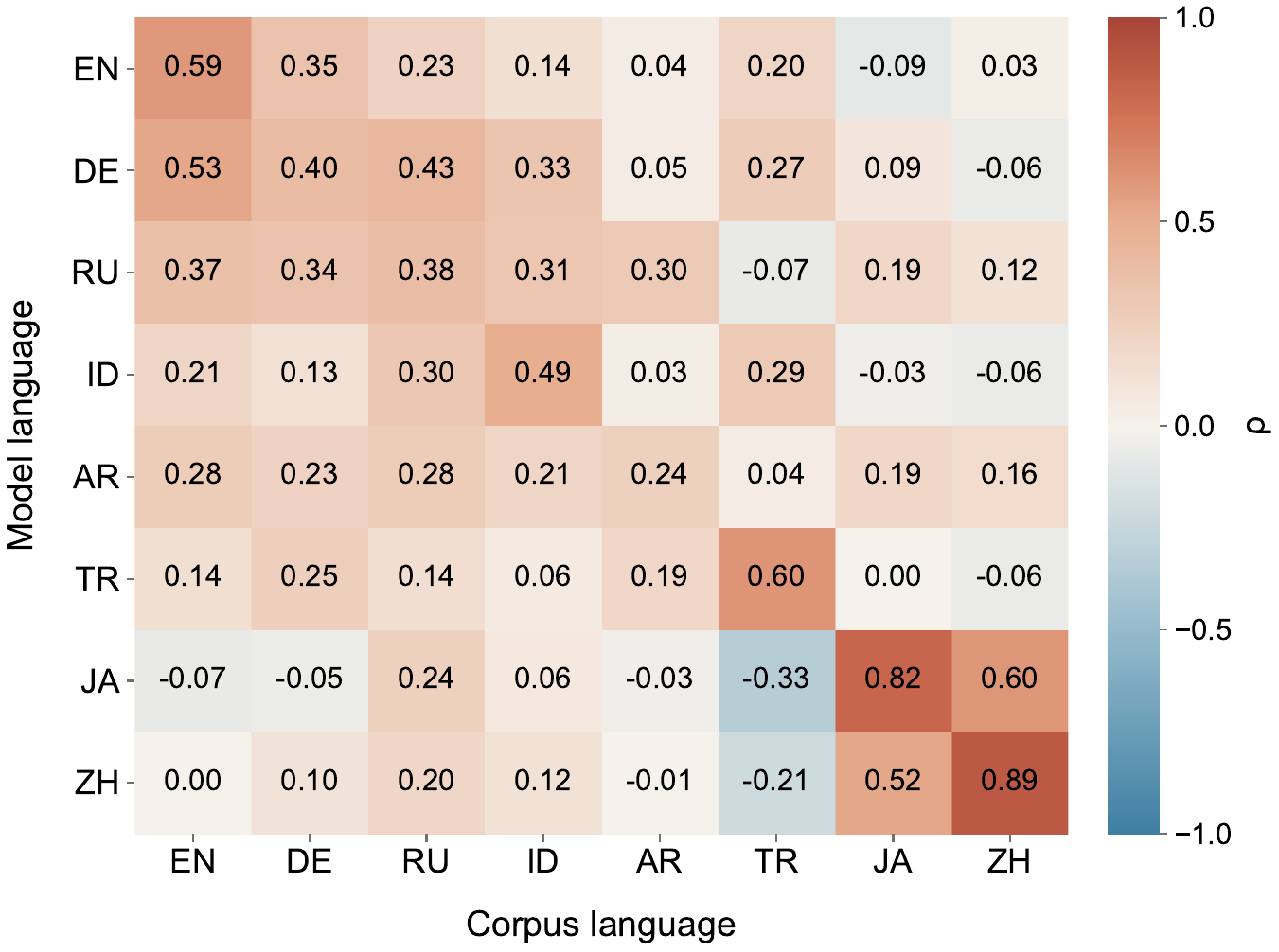}
    \caption{$p_{\text{LLM}}$ vs.\ $p_{\text{corpus}}$}
\end{subfigure}
\hfill
\begin{subfigure}[t]{0.48\textwidth}
    \centering
    \includegraphics[width=\linewidth]{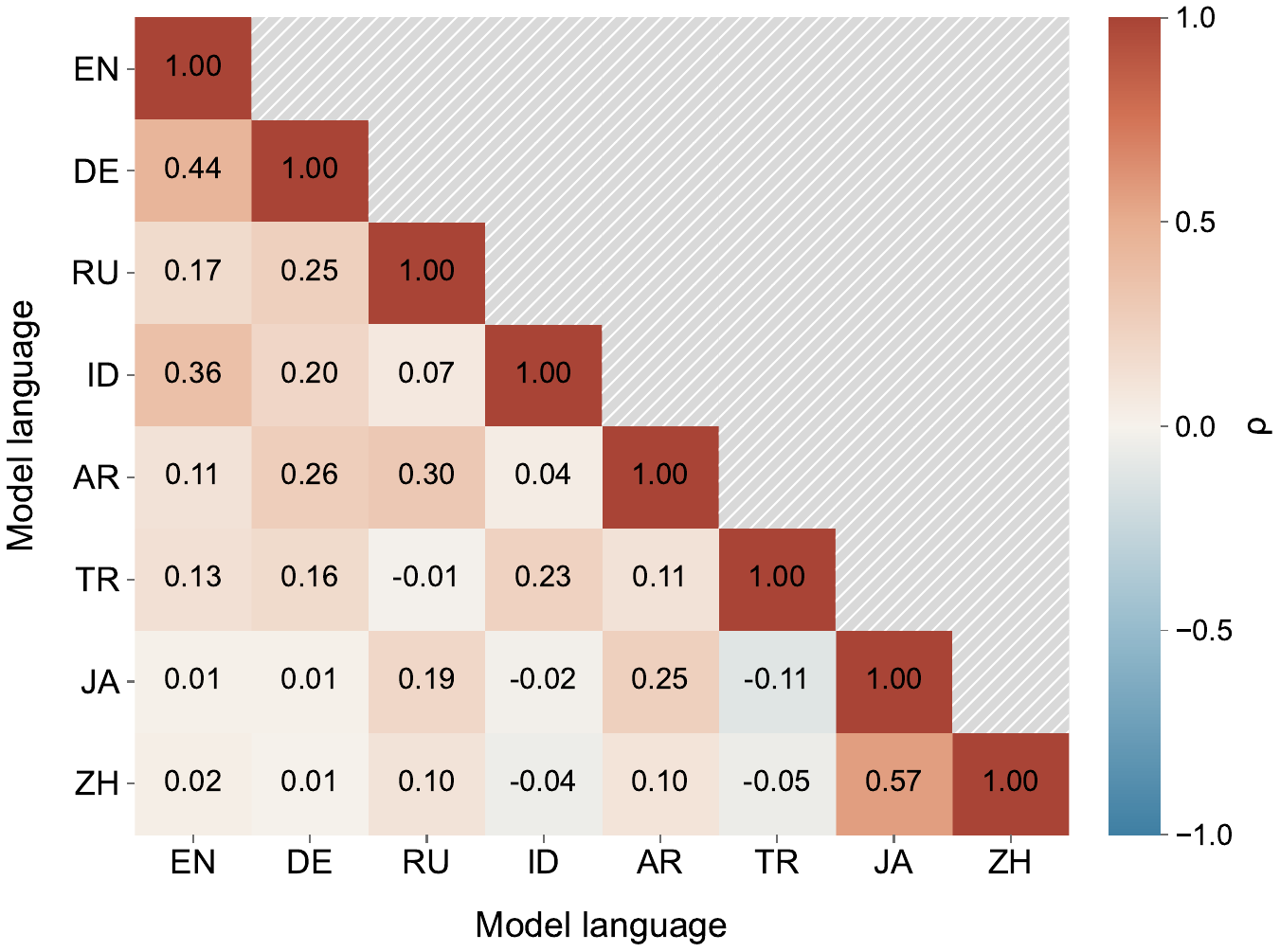}
    \caption{$p_{\text{LLM}}$ vs.\ $p_{\text{LLM}}$}
\end{subfigure}
\caption{Cross-linguistic alignment diagnostics for Llama-3.2-3B.
Left: model--corpus correlations. Right: model--model correlations.
Each cell reports Spearman correlation over shared binomial items.}
\label{fig:appendix-heatmaps-llama3b}
\end{figure*}

\begin{figure*}[t]
\centering
\begin{subfigure}[t]{0.48\textwidth}
    \centering
    \includegraphics[width=\linewidth]{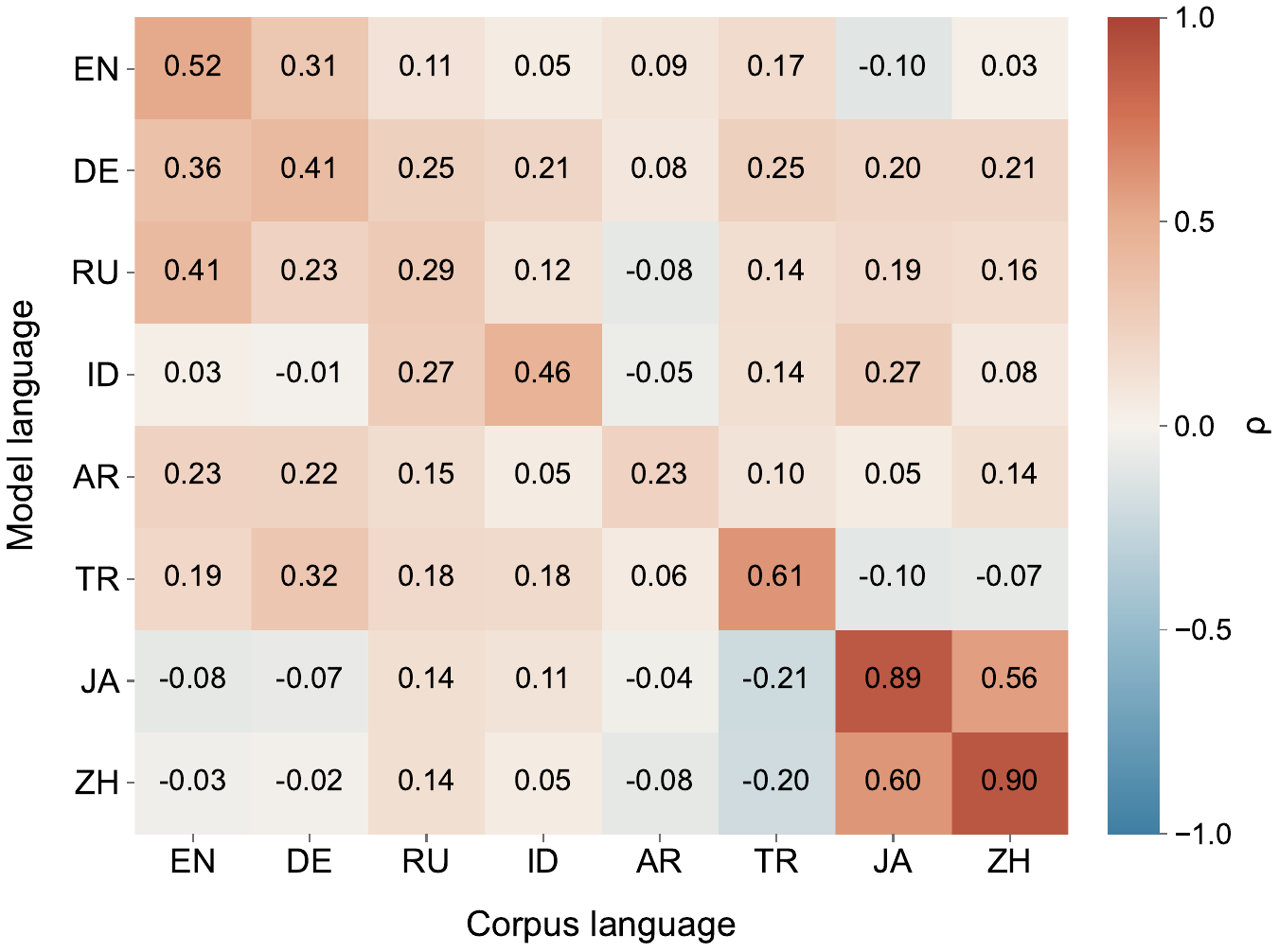}
    \caption{$p_{\text{LLM}}$ vs.\ $p_{\text{corpus}}$}
\end{subfigure}
\hfill
\begin{subfigure}[t]{0.48\textwidth}
    \centering
    \includegraphics[width=\linewidth]{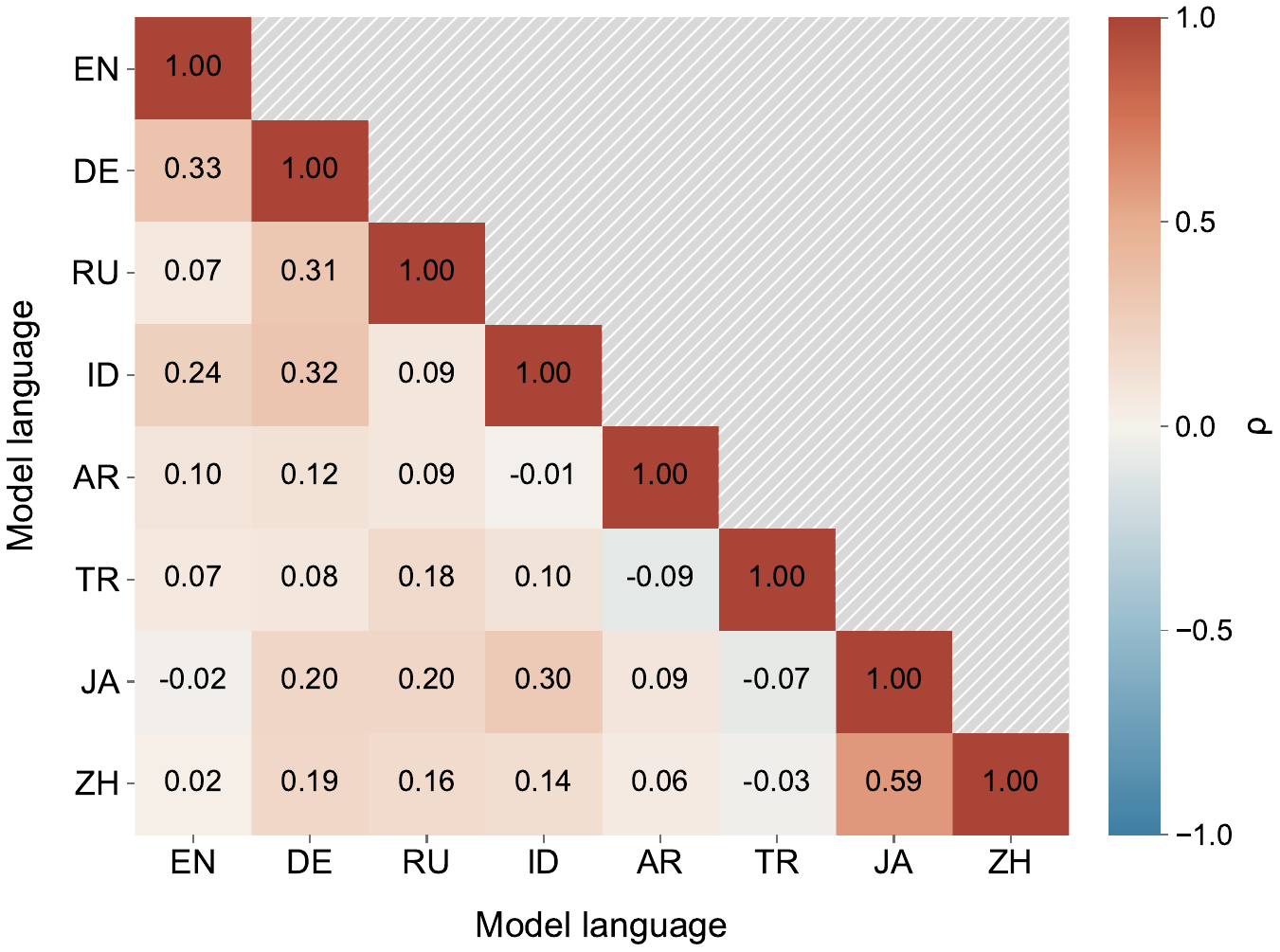}
    \caption{$p_{\text{LLM}}$ vs.\ $p_{\text{LLM}}$}
\end{subfigure}
\caption{Cross-linguistic alignment diagnostics for Gemma-3-4B.
Left: model--corpus correlations. Right: model--model correlations.
Each cell reports Spearman correlation over shared binomial items.}
\label{fig:appendix-heatmaps-gemma4b}
\end{figure*}

\begin{figure*}[t]
\centering
\begin{subfigure}[t]{0.48\textwidth}
    \centering
    \includegraphics[width=\linewidth]{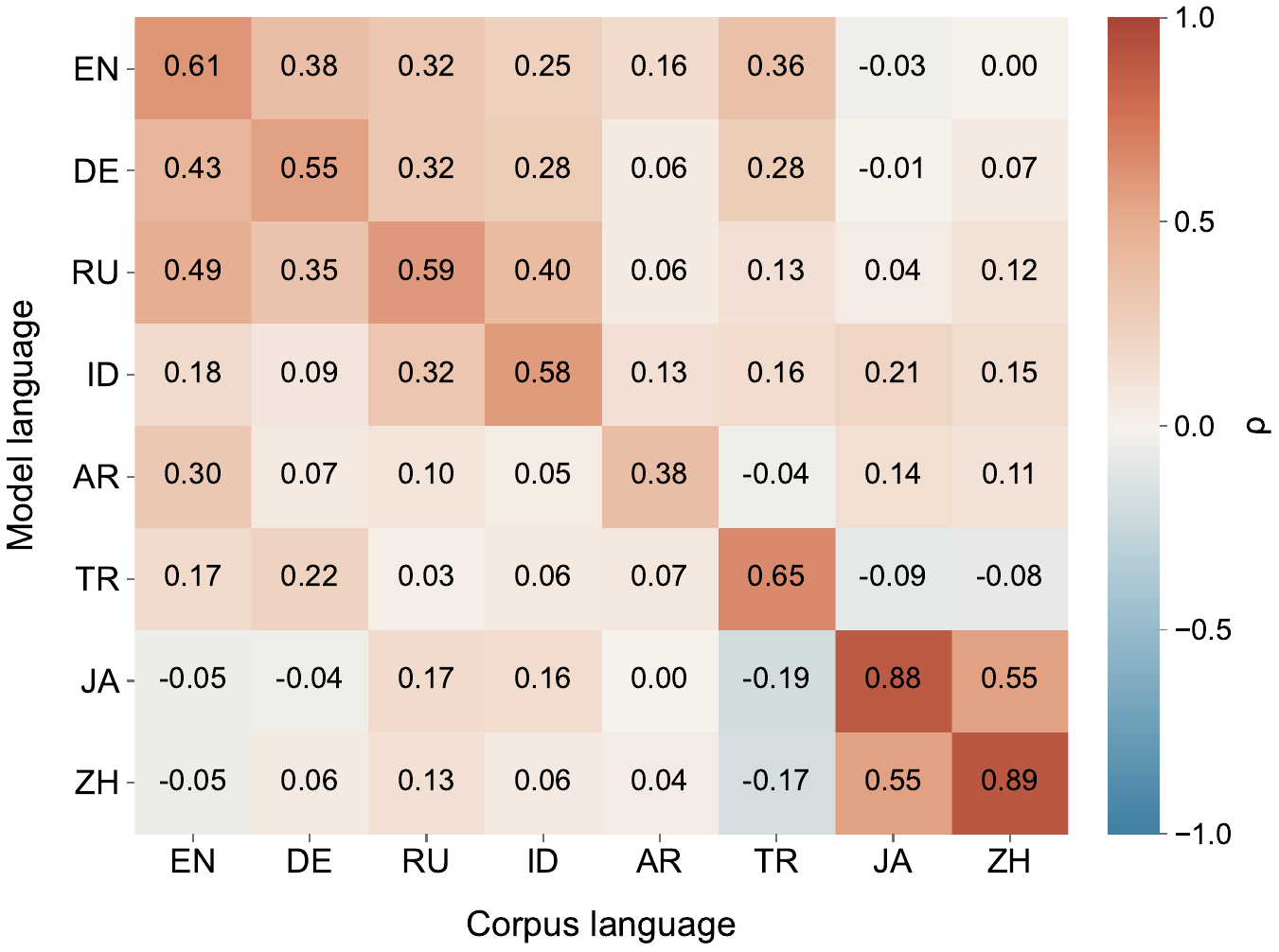}
    \caption{$p_{\text{LLM}}$ vs.\ $p_{\text{corpus}}$}
\end{subfigure}
\hfill
\begin{subfigure}[t]{0.48\textwidth}
    \centering
    \includegraphics[width=\linewidth]{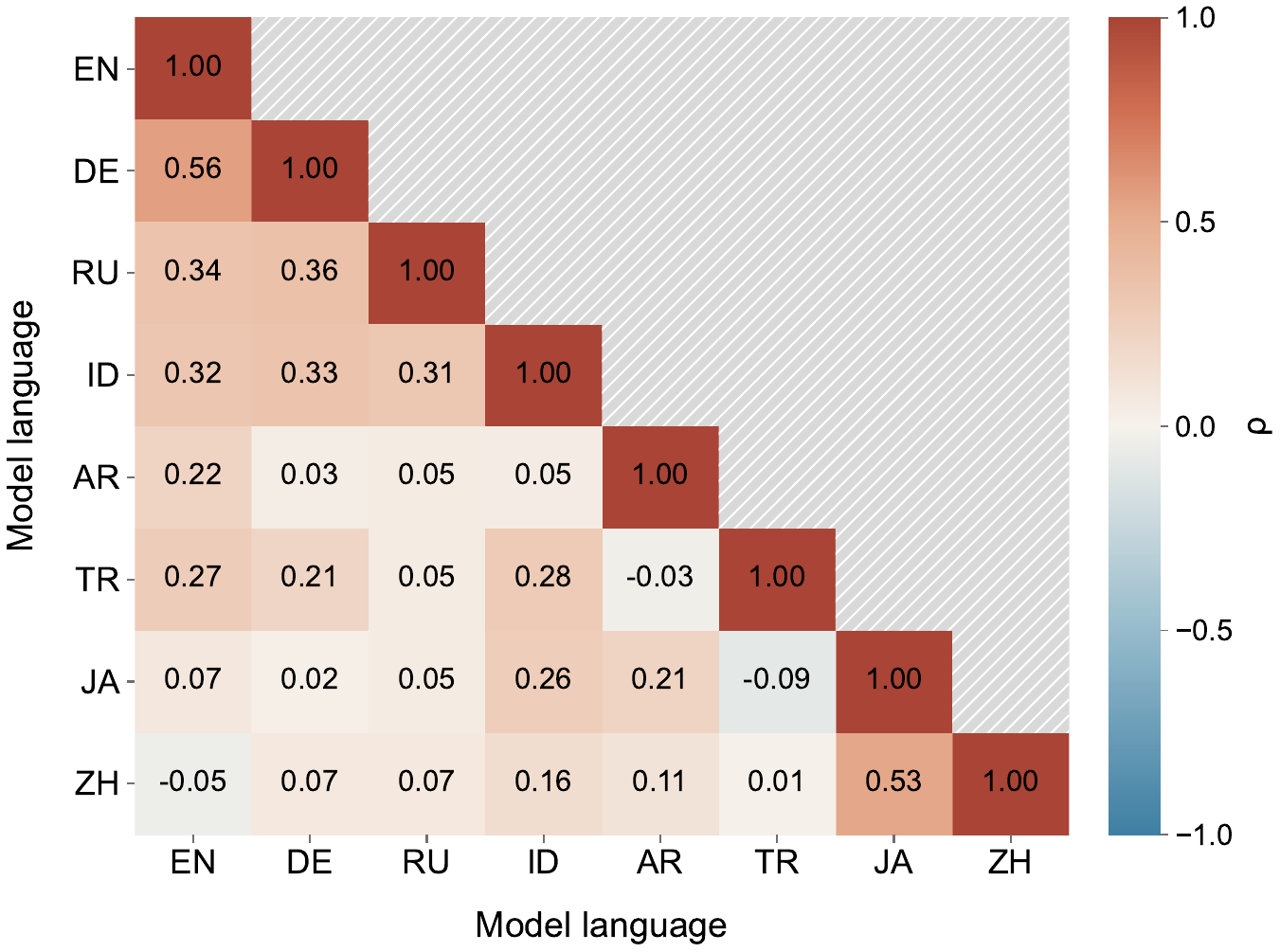}
    \caption{$p_{\text{LLM}}$ vs.\ $p_{\text{LLM}}$}
\end{subfigure}
\caption{Cross-linguistic alignment diagnostics for Gemma-3-12B.
Left: model--corpus correlations. Right: model--model correlations.
Each cell reports Spearman correlation over shared binomial items.}
\label{fig:appendix-heatmaps-gemma12b}
\end{figure*}

Figure~\ref{fig:appendix-crosslingual-heatmaps} shows the corpus--corpus reference and the model--model heatmap for Llama-3.1-8B. Figures~\ref{fig:appendix-heatmaps-qwen4b}--\ref{fig:appendix-heatmaps-gemma12b} provide the corresponding model--corpus and model--model diagnostics for the remaining models.

\subsection{Item-level Alignment}
\label{app:behavioral-scatter}

\begin{figure*}[t]
\centering
\begin{subfigure}[t]{0.48\textwidth}
\includegraphics[width=\linewidth]{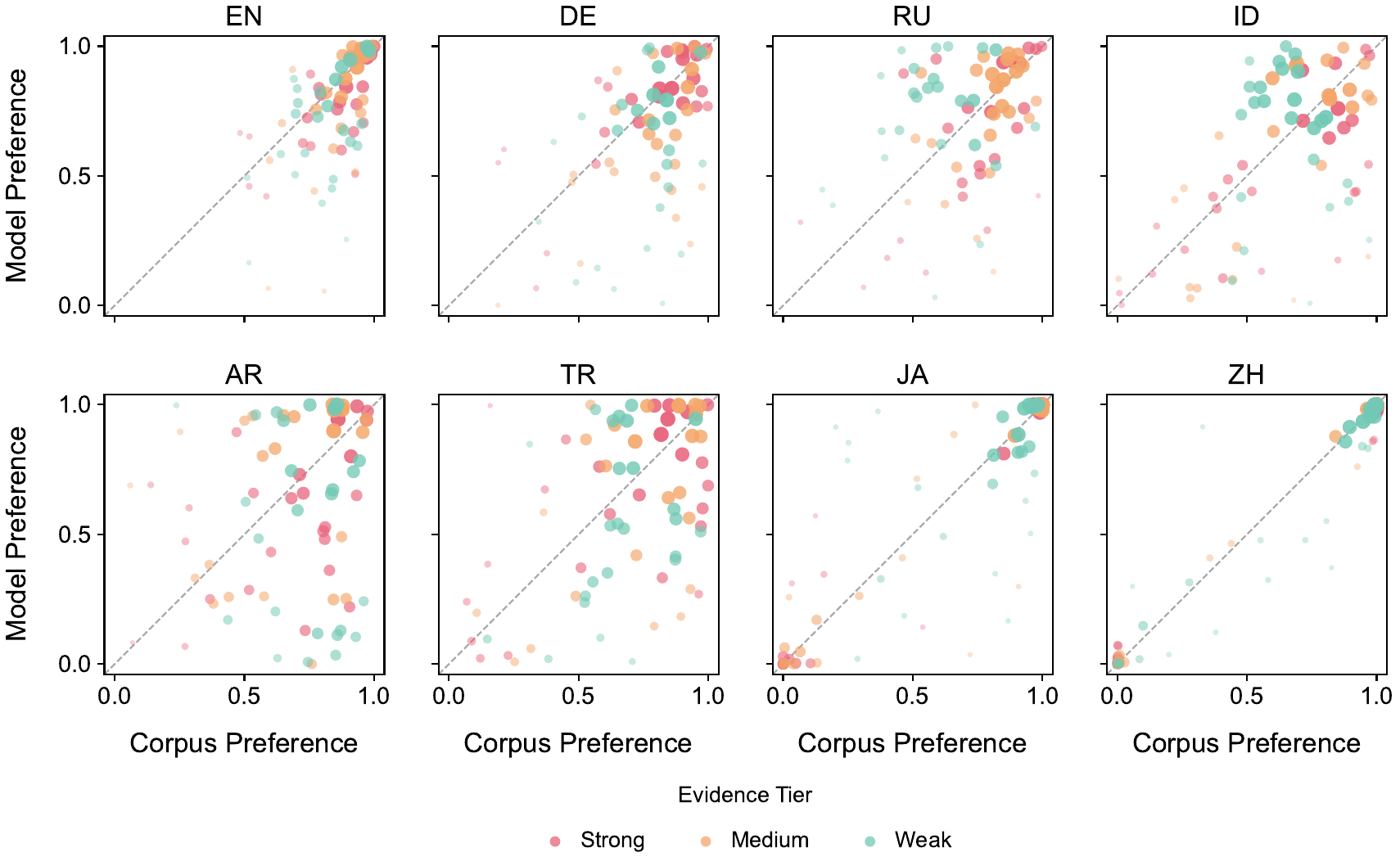}
\caption{Qwen3-4B}
\label{fig:appendix-scatter-qwen4b}
\end{subfigure}
\hfill
\begin{subfigure}[t]{0.48\textwidth}
\includegraphics[width=\linewidth]{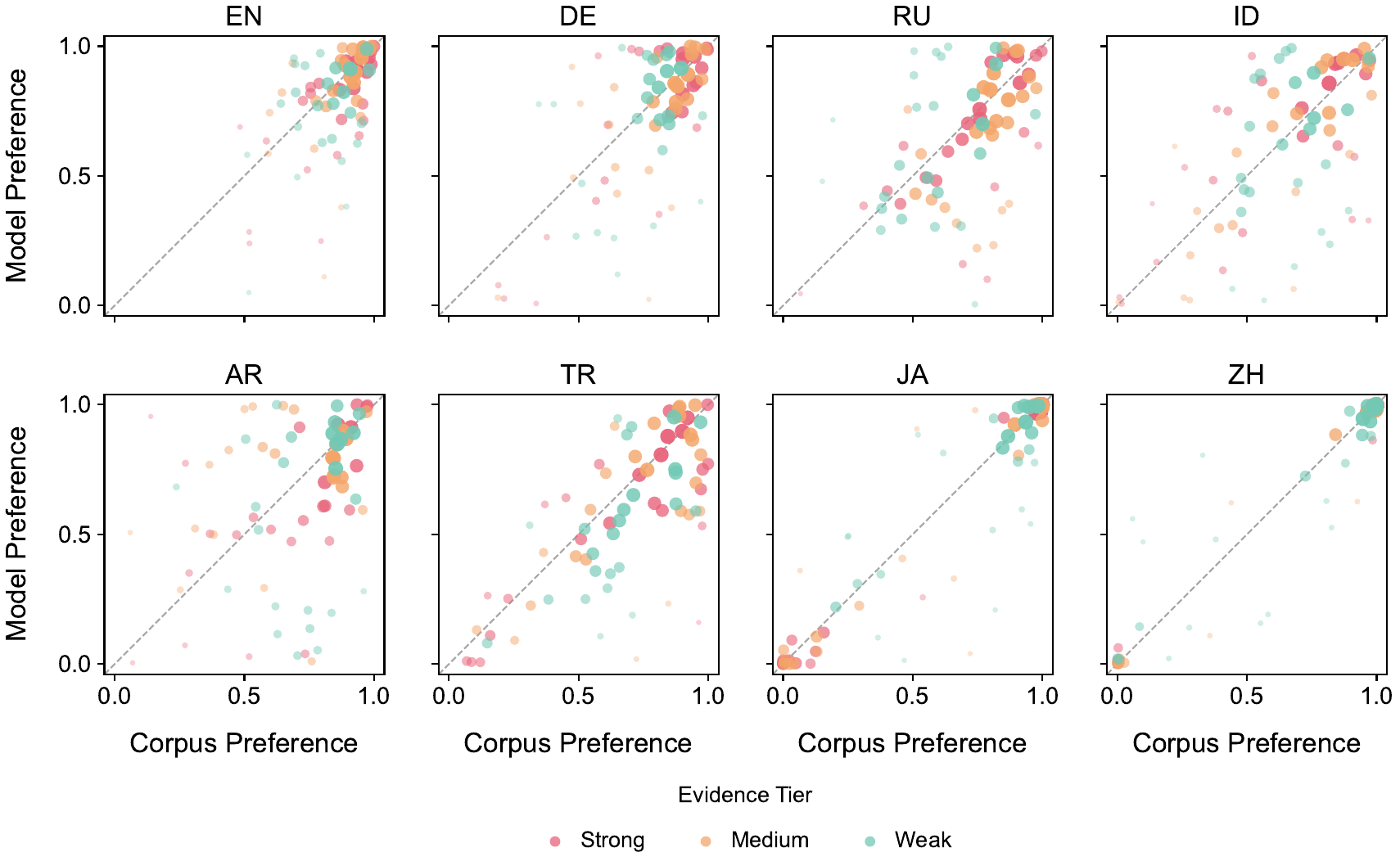}
\caption{Qwen3-14B}
\label{fig:appendix-scatter-qwen14b}
\end{subfigure}
\\
\vspace{1ex}

\begin{subfigure}[t]{0.48\textwidth}
\includegraphics[width=\linewidth]{latex/figures/behavior/llama_8b/scatter_p_llm_vs_p_corpus.pdf}
\caption{Llama-3.1-8B}
\label{fig:appendix-scatter-llama8b}
\end{subfigure}
\hfill
\begin{subfigure}[t]{0.48\textwidth}
\includegraphics[width=\linewidth]{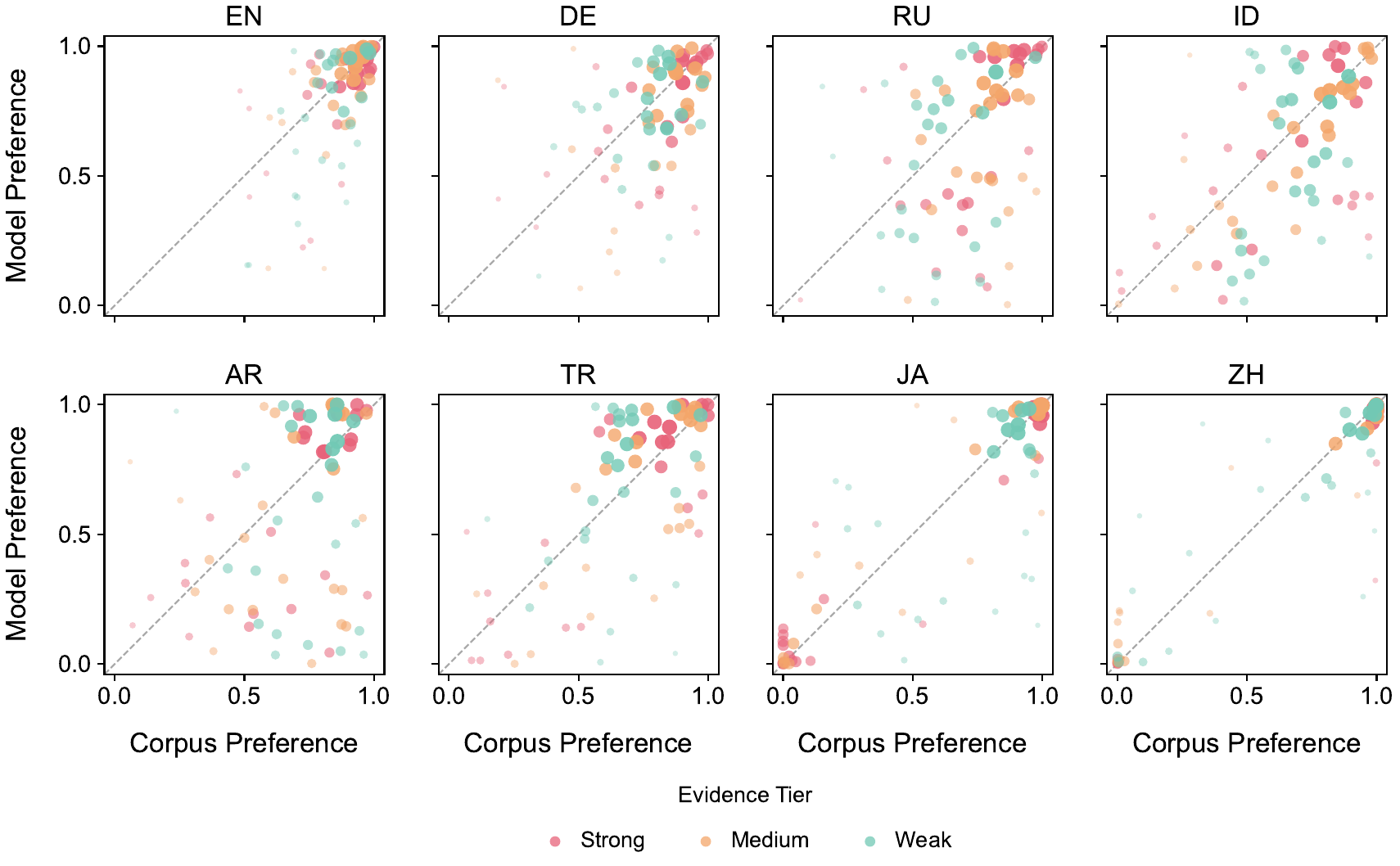}
\caption{Llama-3.2-3B}
\label{fig:appendix-scatter-llama3b}
\end{subfigure}
\\
\vspace{1ex}

\begin{subfigure}[t]{0.48\textwidth}
\includegraphics[width=\linewidth]{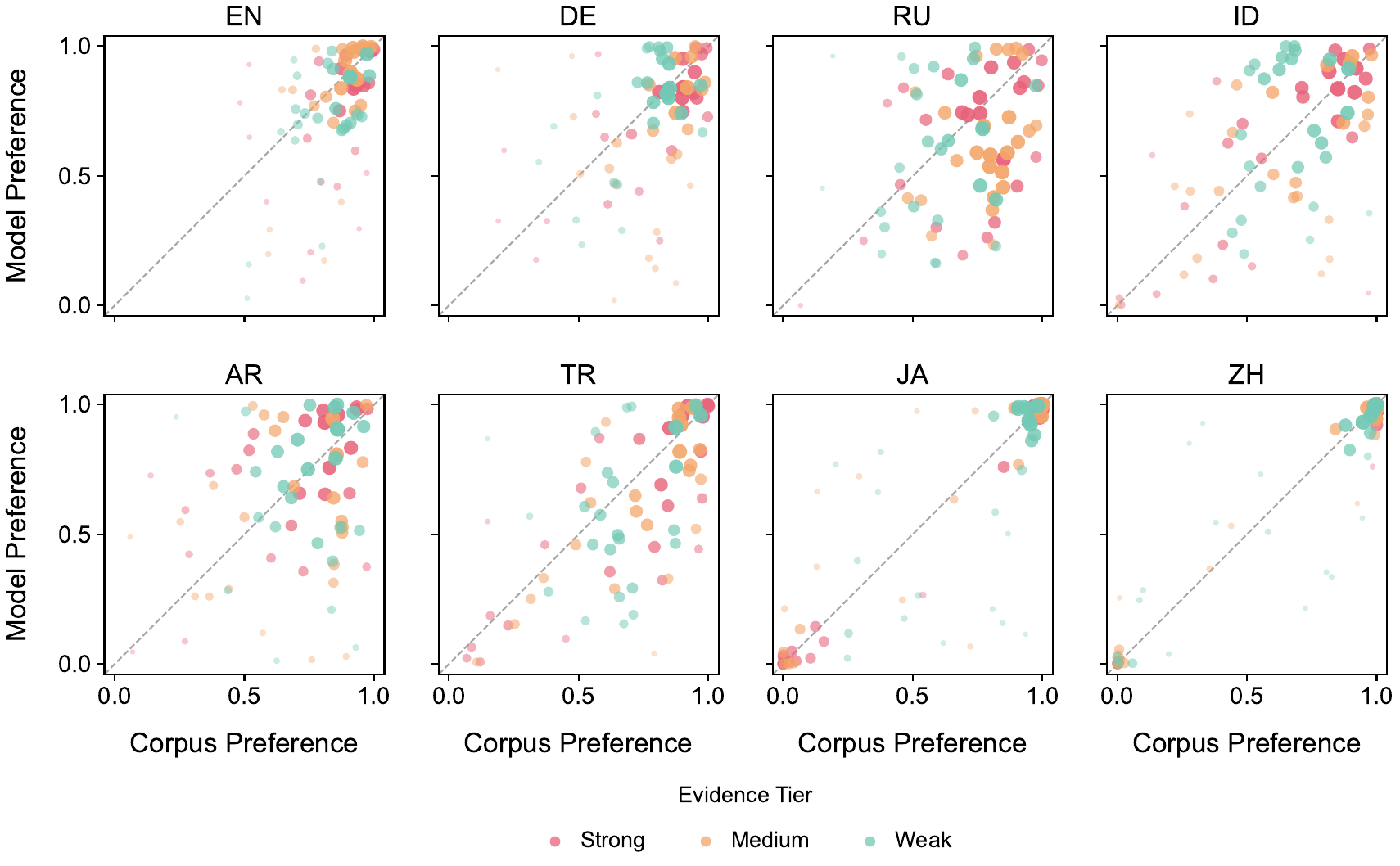}
\caption{Gemma-3-4B}
\label{fig:appendix-scatter-gemma4b}
\end{subfigure}
\hfill
\begin{subfigure}[t]{0.48\textwidth}
\includegraphics[width=\linewidth]{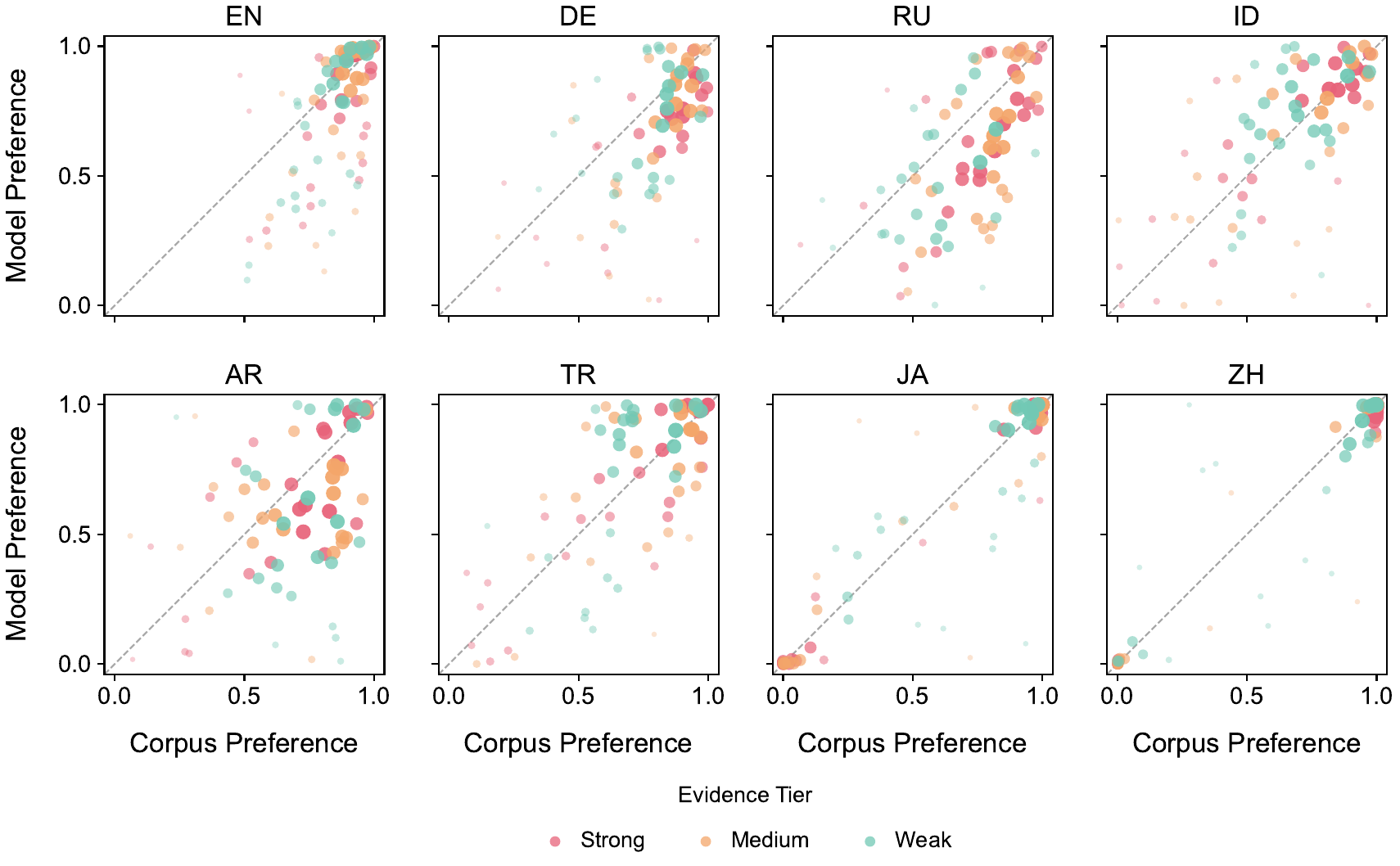}
\caption{Gemma-3-12B}
\label{fig:appendix-scatter-gemma12b}
\end{subfigure}
\caption{Item-level model--corpus alignment for different models across languages.
Points are language-specific binomial items; the diagonal indicates perfect
probability-level agreement.}
\label{fig:appendix-scatter-models}
\end{figure*}

Figure~\ref{fig:appendix-scatter-models} provides item-level scatter plots comparing $p_{\mathrm{LLM}}$ and $p_{\mathrm{corpus}}$ for each language-specific binomial item. Colors indicate corpus evidence tiers based on the combined frequency of the two orders.

\section{Probing Results}
\label{app:probe}

\begin{figure*}[t]
\centering
\includegraphics[
  width=0.78\textwidth,
  height=0.32\textheight,
  keepaspectratio
]{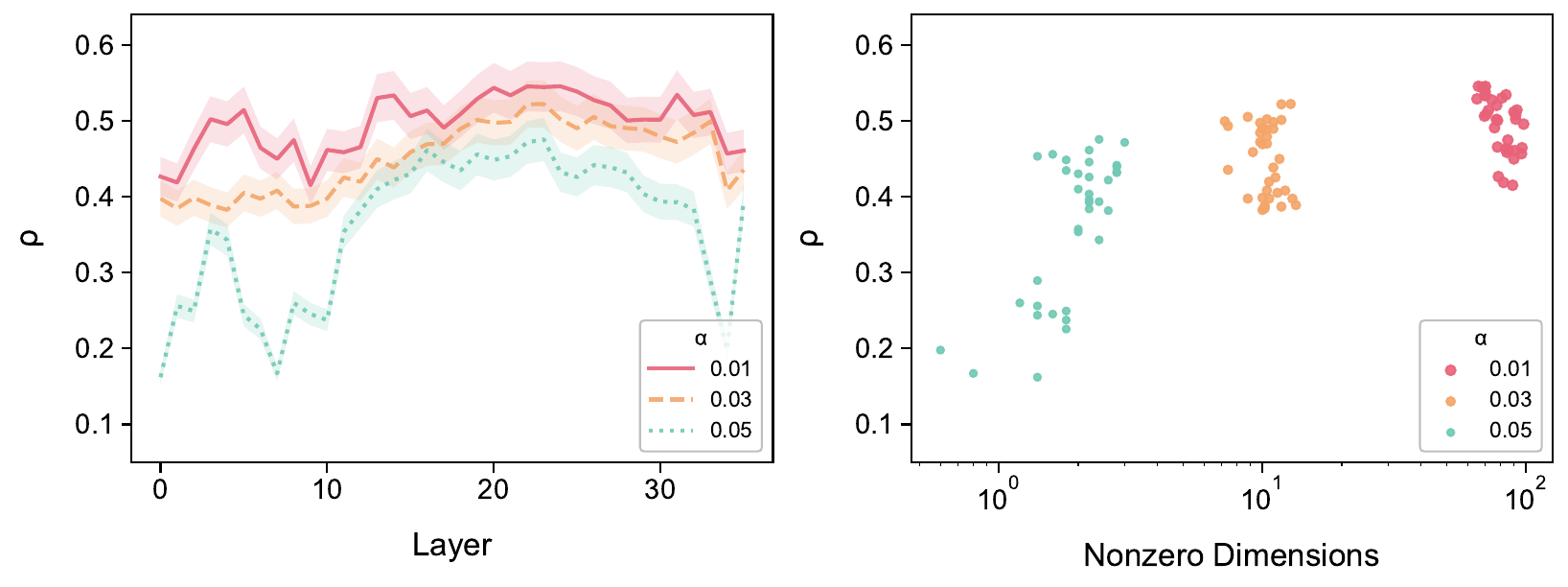}
\caption{
Sparse probing results for corpus-derived preference strength in Qwen3-4B using
mean-pooled binomial representations.
The y-axis shows Spearman correlation between predicted and corpus-derived preference strength; the left x-axis shows decoder layers, and the right x-axis shows the average number of nonzero Lasso coefficients.
Different Lasso regularization strengths are indicated with different colors, line styles (left), and marker sizes (right).
}
\label{fig:probe-lasso-mean}
\end{figure*}

Complementing the last-token results in the main text, Figure~\ref{fig:probe-lasso-mean} reports sparse probing results using mean-pooled binomial representations. Table~\ref{tab:appendix-probing-layer-rho} provides layer-wise Spearman correlations for both last-token and mean-pooled representations, allowing a direct comparison of the two pooling choices across decoder layers.

\begin{table*}[t]
\centering
\small
\setlength{\tabcolsep}{4pt}
\begin{tabular}{rcc|rcc|rcc|rcc}
\toprule
\textbf{Layer} & \textbf{Last $\rho$} & \textbf{Mean $\rho$} &
\textbf{Layer} & \textbf{Last $\rho$} & \textbf{Mean $\rho$} &
\textbf{Layer} & \textbf{Last $\rho$} & \textbf{Mean $\rho$} &
\textbf{Layer} & \textbf{Last $\rho$} & \textbf{Mean $\rho$} \\
\midrule
0 & .367 & .397 & 9  & .409 & .388          & 18 & .478 & .489 & 27 & .485 & .493 \\
1 & .358 & .385 & 10 & .409 & .397          & 19 & .481 & .501 & 28 & .491 & .490 \\
2 & .404 & .399 & 11 & .439 & .425          & 20 & .491 & .497 & 29 & .488 & .488 \\
3 & .394 & .389 & 12 & .485 & .420          & 21 & .497 & .498 & 30 & .477 & .479 \\
4 & .424 & .383 & 13 & .499 & .450          & 22 & .500 & .522 & 31 & .486 & .472 \\
5 & .404 & .405 & 14 & \textbf{.522} & .438 & 23 & .511 & \textbf{.522} & 32 & .464 & .484 \\
6 & .427 & .397 & 15 & .504 & .459          & 24 & .469 & .502 & 33 & .460 & .499 \\
7 & .401 & .408 & 16 & .475 & .469          & 25 & .493 & .490 & 34 & .442 & .408 \\
8 & .403 & .387 & 17 & .494 & .470          & 26 & .498 & .505 & 35 & .414 & .435 \\
\bottomrule
\end{tabular}
\caption{Layer-wise probing results for Qwen3-4B.
Values are mean Spearman $\rho$ across five folds.
Probes are trained using either the last-token representation or the mean-pooled representation.}
\label{tab:appendix-probing-layer-rho}
\end{table*}

\section{Steering Across Scales}
\label{app:graded-steering}

\begin{figure*}[t]
\centering
\includegraphics[width=.7\textwidth]{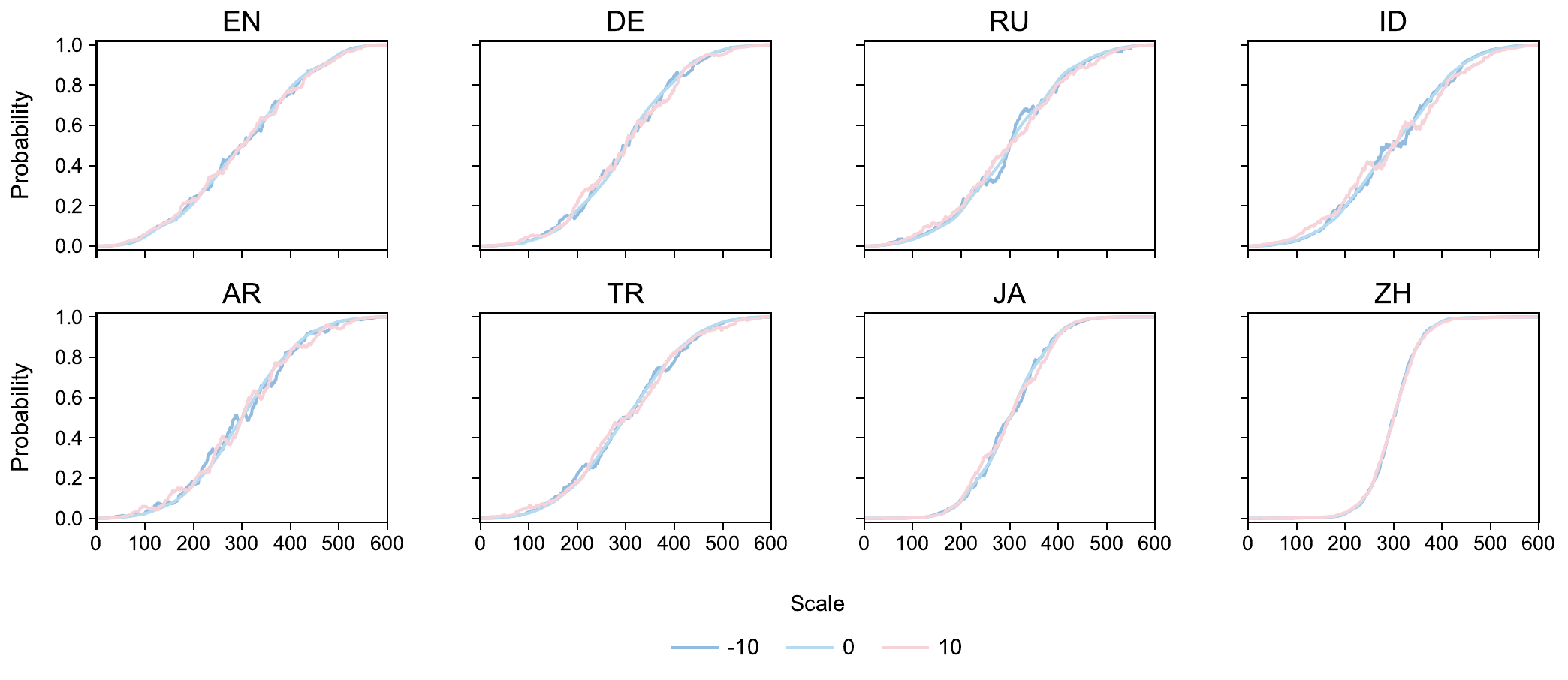}
\includegraphics[width=.7\textwidth]{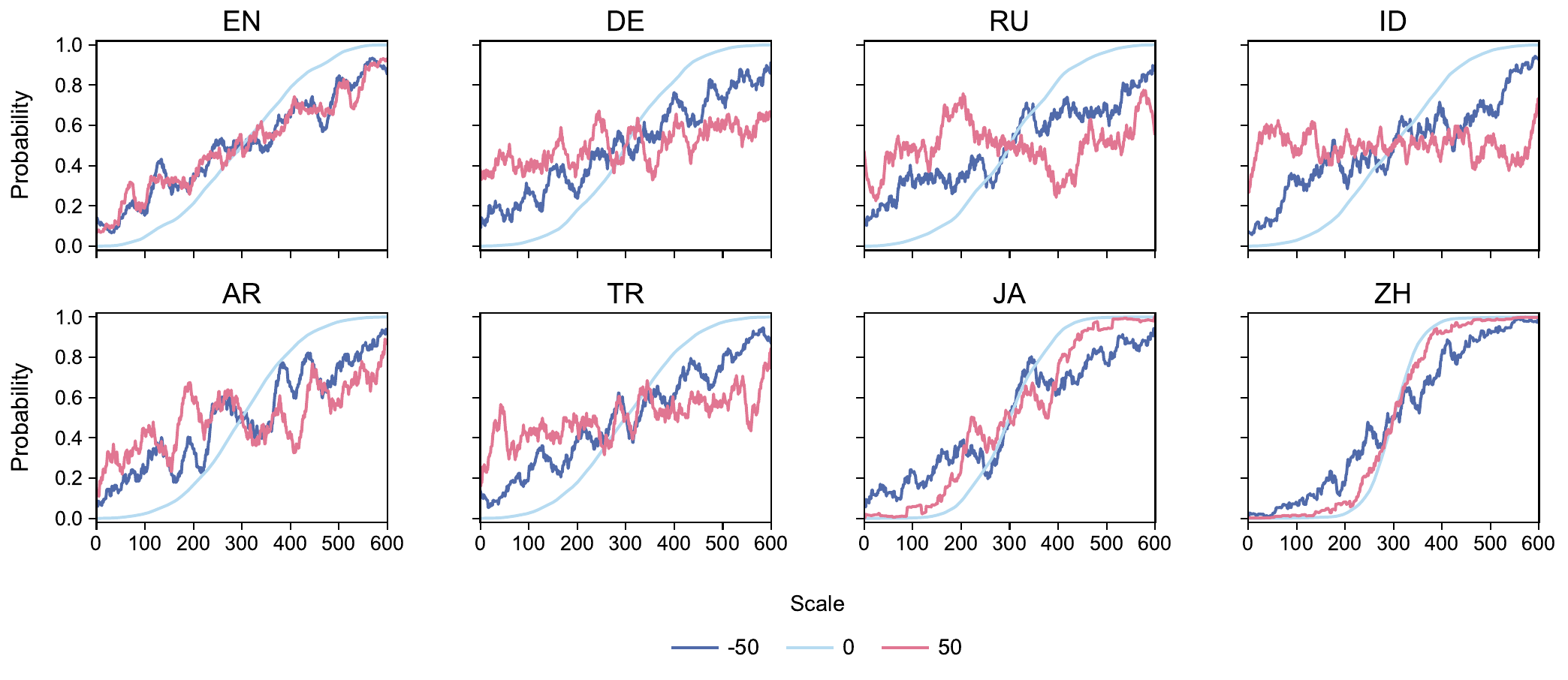}
\caption{
Effects of steering on Qwen3-4B binomial order probabilities using the last-token steering vector at layer 14.
The upper panel compares $\lambda=-10,0,+10$, and the lower panel compares $\lambda=-50,0,+50$, with $\lambda=0$ denoting the original baseline.
Each language panel sorts binomial orders by their original model probability.
}
\label{fig:appendix-steering-10}
\end{figure*}

\begin{figure*}[t]
\centering
\includegraphics[width=.7\textwidth]{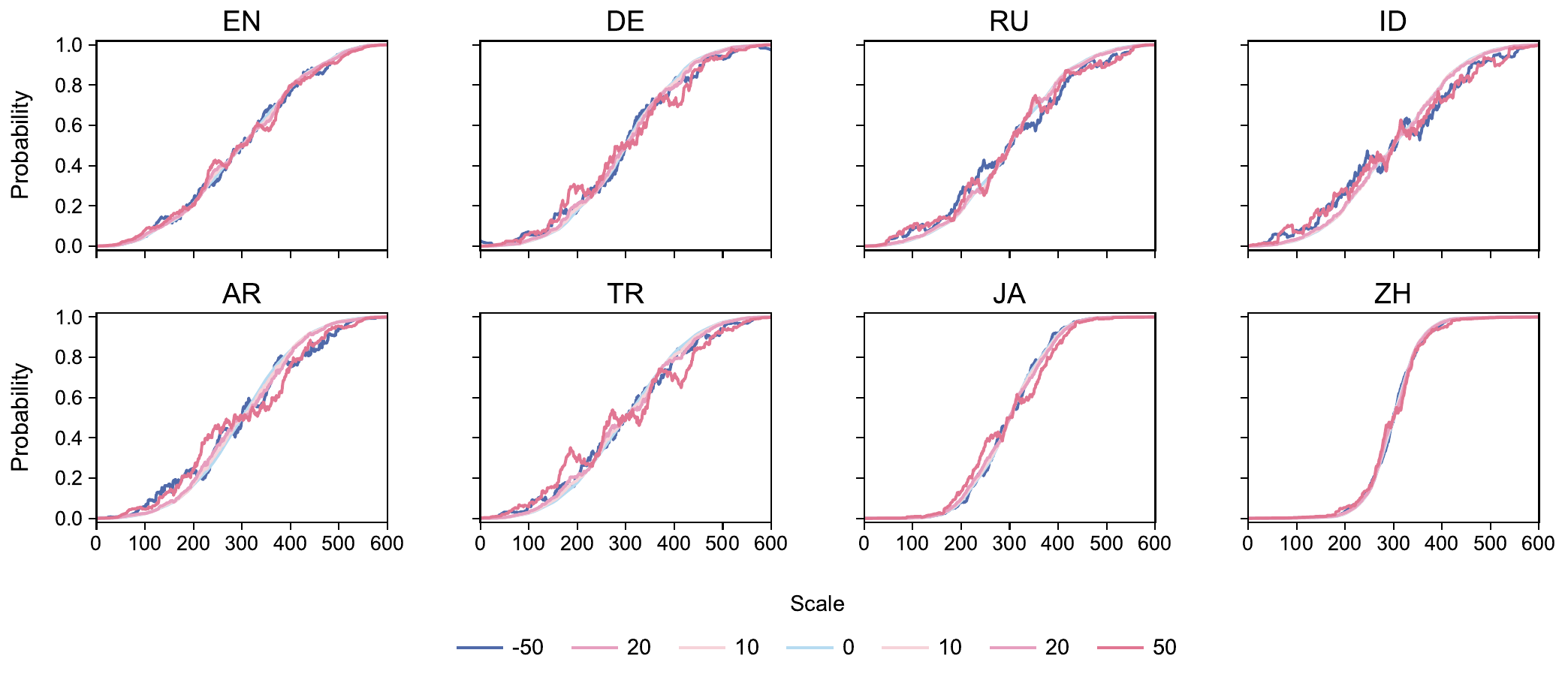}
\caption{
Effects of steering on Qwen3-4B binomial order probabilities using the mean-pooled steering vector at layer 23.
Curves compare $\lambda \in \{-50,-20,-10,0,+10,+20,+50\}$, with $\lambda=0$ denoting the original baseline.
Each language panel sorts binomial orders by their original model probability.
}
\label{fig:appendix-steering-mean23-all}
\end{figure*}

Figure~\ref{fig:appendix-steering-10} extends the last-token steering analysis at layer 14 to additional intervention scales. As a robustness check, Figure~\ref{fig:appendix-steering-mean23-all} shows the corresponding mean-pooled steering results at layer 23.

%% file: acl_latex.bbl
\begin{thebibliography}{48}
\providecommand{\natexlab}[1]{#1}

\bibitem[{Ahuja et~al.(2023)Ahuja, Diddee, Hada, Ochieng, Ramesh, Jain, Nambi, Ganu, Segal, Axmed, Bali, and Sitaram}]{ahuja2023megamultilingualevaluationgenerative}
Kabir Ahuja, Harshita Diddee, Rishav Hada, Millicent Ochieng, Krithika Ramesh, Prachi Jain, Akshay Nambi, Tanuja Ganu, Sameer Segal, Maxamed Axmed, Kalika Bali, and Sunayana Sitaram. 2023.
\newblock \href {https://arxiv.org/abs/2303.12528} {Mega: Multilingual evaluation of generative ai}.
\newblock \emph{Preprint}, arXiv:2303.12528.

\bibitem[{Alain and Bengio(2018)}]{alain2018understandingintermediatelayersusing}
Guillaume Alain and Yoshua Bengio. 2018.
\newblock \href {https://arxiv.org/abs/1610.01644} {Understanding intermediate layers using linear classifier probes}.
\newblock \emph{Preprint}, arXiv:1610.01644.

\bibitem[{Belinkov et~al.(2017)Belinkov, Durrani, Dalvi, Sajjad, and Glass}]{belinkov-etal-2017-neural}
Yonatan Belinkov, Nadir Durrani, Fahim Dalvi, Hassan Sajjad, and James~R. Glass. 2017.
\newblock \href {https://arxiv.org/abs/1704.03471} {What do neural machine translation models learn about morphology?}
\newblock \emph{CoRR}, abs/1704.03471.

\bibitem[{Benor and Levy(2006)}]{benor2006}
Sarah~Bunin Benor and Roger Levy. 2006.
\newblock \href {http://www.jstor.org/stable/4490157} {The chicken or the egg? a probabilistic analysis of english binomials}.
\newblock \emph{Language}, 82(2):233--278.

\bibitem[{Cooper and Ross(1975)}]{cooperross1975worldorder}
William~E. Cooper and John~Robert Ross. 1975.
\newblock \href {https://www.academia.edu/28663645/World_order_by_William_Cooper_and_Haj_Ross} {World order}.
\newblock In Robin~E. Grossman, L.~James San, and Timothy~J. Vance, editors, \emph{Papers from the Parasession on Functionalism}, pages 63--111. Chicago Linguistic Society, Chicago, IL.

\bibitem[{De~Luca~Fornaciari et~al.(2024)De~Luca~Fornaciari, Altuna, Gonzalez-Dios, and Melero}]{de-luca-fornaciari-etal-2024-hard}
Francesca De~Luca~Fornaciari, Bego{\~n}a Altuna, Itziar Gonzalez-Dios, and Maite Melero. 2024.
\newblock \href {https://doi.org/10.18653/v1/2024.figlang-1.5} {A hard nut to crack: Idiom detection with conversational large language models}.
\newblock In \emph{Proceedings of the 4th Workshop on Figurative Language Processing (FigLang 2024)}, pages 35--44, Mexico City, Mexico (Hybrid). Association for Computational Linguistics.

\bibitem[{Espinosa-Anke et~al.(2021)Espinosa-Anke, Codina-Filb{\`a}, and Wanner}]{espinosa-anke-etal-2021-evaluating}
Luis Espinosa-Anke, Joan Codina-Filb{\`a}, and Leo Wanner. 2021.
\newblock \href {https://doi.org/10.18653/v1/2021.eacl-main.120} {Evaluating language models for the retrieval and categorization of lexical collocations}.
\newblock In \emph{Proceedings of the 16th Conference of the European Chapter of the Association for Computational Linguistics: Main Volume}, pages 1406--1417, Online. Association for Computational Linguistics.

\bibitem[{Grattafiori et~al.(2024)Grattafiori, Dubey, Jauhri, Pandey, Kadian, Al-Dahle, Letman, Mathur, Schelten, Vaughan, Yang, Fan, Goyal, Hartshorn, Yang, Mitra, Sravankumar, Korenev, Hinsvark, Rao, Zhang, Rodriguez, Gregerson, Spataru, Roziere, Biron, Tang, Chern, Caucheteux, Nayak, Bi, Marra, McConnell, Keller, Touret, Wu, Wong, Ferrer, Nikolaidis, Allonsius, Song, Pintz, Livshits, Wyatt, Esiobu, Choudhary, Mahajan, Garcia-Olano, Perino, Hupkes, Lakomkin, AlBadawy, Lobanova, Dinan, Smith, Radenovic, Guzmán, Zhang, Synnaeve, Lee, Anderson, Thattai, Nail, Mialon, Pang, Cucurell, Nguyen, Korevaar, Xu, Touvron, Zarov, Ibarra, Kloumann, Misra, Evtimov, Zhang, Copet, Lee, Geffert, Vranes, Park, Mahadeokar, Shah, van~der Linde, Billock, Hong, Lee, Fu, Chi, Huang, Liu, Wang, Yu, Bitton, Spisak, Park, Rocca, Johnstun, Saxe, Jia, Alwala, Prasad, Upasani, Plawiak, Li, Heafield, Stone, El-Arini, Iyer, Malik, Chiu, Bhalla, Lakhotia, Rantala-Yeary, van~der Maaten, Chen, Tan, Jenkins, Martin, Madaan, Malo, Blecher,
  Landzaat, de~Oliveira, Muzzi, Pasupuleti, Singh, Paluri, Kardas, Tsimpoukelli, Oldham, Rita, Pavlova, Kambadur, Lewis, Si, Singh, Hassan, Goyal, Torabi, Bashlykov, Bogoychev, Chatterji, Zhang, Duchenne, Çelebi, Alrassy, Zhang, Li, Vasic, Weng, Bhargava, Dubal, Krishnan, Koura, Xu, He, Dong, Srinivasan, Ganapathy, Calderer, Cabral, Stojnic, Raileanu, Maheswari, Girdhar, Patel, Sauvestre, Polidoro, Sumbaly, Taylor, Silva, Hou, Wang, Hosseini, Chennabasappa, Singh, Bell, Kim, Edunov, Nie, Narang, Raparthy, Shen, Wan, Bhosale, Zhang, Vandenhende, Batra, Whitman, Sootla, Collot, Gururangan, Borodinsky, Herman, Fowler, Sheasha, Georgiou, Scialom, Speckbacher, Mihaylov, Xiao, Karn, Goswami, Gupta, Ramanathan, Kerkez, Gonguet, Do, Vogeti, Albiero, Petrovic, Chu, Xiong, Fu, Meers, Martinet, Wang, Wang, Tan, Xia, Xie, Jia, Wang, Goldschlag, Gaur, Babaei, Wen, Song, Zhang, Li, Mao, Coudert, Yan, Chen, Papakipos, Singh, Srivastava, Jain, Kelsey, Shajnfeld, Gangidi, Victoria, Goldstand, Menon, Sharma, Boesenberg,
  Baevski, Feinstein, Kallet, Sangani, Teo, Yunus, Lupu, Alvarado, Caples, Gu, Ho, Poulton, Ryan, Ramchandani, Dong, Franco, Goyal, Saraf, Chowdhury, Gabriel, Bharambe, Eisenman, Yazdan, James, Maurer, Leonhardi, Huang, Loyd, Paola, Paranjape, Liu, Wu, Ni, Hancock, Wasti, Spence, Stojkovic, Gamido, Montalvo, Parker, Burton, Mejia, Liu, Wang, Kim, Zhou, Hu, Chu, Cai, Tindal, Feichtenhofer, Gao, Civin, Beaty, Kreymer, Li, Adkins, Xu, Testuggine, David, Parikh, Liskovich, Foss, Wang, Le, Holland, Dowling, Jamil, Montgomery, Presani, Hahn, Wood, Le, Brinkman, Arcaute, Dunbar, Smothers, Sun, Kreuk, Tian, Kokkinos, Ozgenel, Caggioni, Kanayet, Seide, Florez, Schwarz, Badeer, Swee, Halpern, Herman, Sizov, Guangyi, Zhang, Lakshminarayanan, Inan, Shojanazeri, Zou, Wang, Zha, Habeeb, Rudolph, Suk, Aspegren, Goldman, Zhan, Damlaj, Molybog, Tufanov, Leontiadis, Veliche, Gat, Weissman, Geboski, Kohli, Lam, Asher, Gaya, Marcus, Tang, Chan, Zhen, Reizenstein, Teboul, Zhong, Jin, Yang, Cummings, Carvill, Shepard, McPhie,
  Torres, Ginsburg, Wang, Wu, U, Saxena, Khandelwal, Zand, Matosich, Veeraraghavan, Michelena, Li, Jagadeesh, Huang, Chawla, Huang, Chen, Garg, A, Silva, Bell, Zhang, Guo, Yu, Moshkovich, Wehrstedt, Khabsa, Avalani, Bhatt, Mankus, Hasson, Lennie, Reso, Groshev, Naumov, Lathi, Keneally, Liu, Seltzer, Valko, Restrepo, Patel, Vyatskov, Samvelyan, Clark, Macey, Wang, Hermoso, Metanat, Rastegari, Bansal, Santhanam, Parks, White, Bawa, Singhal, Egebo, Usunier, Mehta, Laptev, Dong, Cheng, Chernoguz, Hart, Salpekar, Kalinli, Kent, Parekh, Saab, Balaji, Rittner, Bontrager, Roux, Dollar, Zvyagina, Ratanchandani, Yuvraj, Liang, Alao, Rodriguez, Ayub, Murthy, Nayani, Mitra, Parthasarathy, Li, Hogan, Battey, Wang, Howes, Rinott, Mehta, Siby, Bondu, Datta, Chugh, Hunt, Dhillon, Sidorov, Pan, Mahajan, Verma, Yamamoto, Ramaswamy, Lindsay, Lindsay, Feng, Lin, Zha, Patil, Shankar, Zhang, Zhang, Wang, Agarwal, Sajuyigbe, Chintala, Max, Chen, Kehoe, Satterfield, Govindaprasad, Gupta, Deng, Cho, Virk, Subramanian, Choudhury,
  Goldman, Remez, Glaser, Best, Koehler, Robinson, Li, Zhang, Matthews, Chou, Shaked, Vontimitta, Ajayi, Montanez, Mohan, Kumar, Mangla, Ionescu, Poenaru, Mihailescu, Ivanov, Li, Wang, Jiang, Bouaziz, Constable, Tang, Wu, Wang, Wu, Gao, Kleinman, Chen, Hu, Jia, Qi, Li, Zhang, Zhang, Adi, Nam, Yu, Wang, Zhao, Hao, Qian, Li, He, Rait, DeVito, Rosnbrick, Wen, Yang, Zhao, and Ma}]{grattafiori2024llama3herdmodels}
Aaron Grattafiori, Abhimanyu Dubey, Abhinav Jauhri, Abhinav Pandey, Abhishek Kadian, Ahmad Al-Dahle, Aiesha Letman, Akhil Mathur, Alan Schelten, Alex Vaughan, Amy Yang, Angela Fan, Anirudh Goyal, Anthony Hartshorn, Aobo Yang, Archi Mitra, Archie Sravankumar, Artem Korenev, Arthur Hinsvark, and 542 others. 2024.
\newblock \href {https://arxiv.org/abs/2407.21783} {The llama 3 herd of models}.
\newblock \emph{Preprint}, arXiv:2407.21783.

\bibitem[{Guo et~al.(2025)Guo, Shang, and Clavel}]{guo2025benchmarkinglinguisticdiversitylarge}
Yanzhu Guo, Guokan Shang, and Chloé Clavel. 2025.
\newblock \href {https://arxiv.org/abs/2412.10271} {Benchmarking linguistic diversity of large language models}.
\newblock \emph{Preprint}, arXiv:2412.10271.

\bibitem[{Gurnee et~al.(2023)Gurnee, Nanda, Pauly, Harvey, Troitskii, and Bertsimas}]{gurnee2023finding}
Wes Gurnee, Neel Nanda, Matthew Pauly, Katherine Harvey, Dmitrii Troitskii, and Dimitris Bertsimas. 2023.
\newblock \href {https://arxiv.org/abs/2305.01610} {Finding neurons in a haystack: Case studies with sparse probing}.
\newblock \emph{ArXiv preprint}, abs/2305.01610.

\bibitem[{Hewitt and Manning(2019)}]{hewitt-manning-2019-structural}
John Hewitt and Christopher~D. Manning. 2019.
\newblock \href {https://doi.org/10.18653/v1/N19-1419} {{A} structural probe for finding syntax in word representations}.
\newblock In \emph{Proceedings of the 2019 Conference of the North {A}merican Chapter of the Association for Computational Linguistics: Human Language Technologies, Volume 1 (Long and Short Papers)}, pages 4129--4138, Minneapolis, Minnesota. Association for Computational Linguistics.

\bibitem[{Houghton et~al.(2025)Houghton, Sagae, and Morgan}]{houghton-etal-2025-role}
Zachary~Nicholas Houghton, Kenji Sagae, and Emily Morgan. 2025.
\newblock \href {https://doi.org/10.18653/v1/2025.acl-short.55} {The role of abstract representations and observed preferences in the ordering of binomials in large language models}.
\newblock In \emph{Proceedings of the 63rd Annual Meeting of the Association for Computational Linguistics (Volume 2: Short Papers)}, pages 695--702, Vienna, Austria. Association for Computational Linguistics.

\bibitem[{Hu et~al.(2024)Hu, Mahowald, Lupyan, Ivanova, and Levy}]{Hu_2024}
Jennifer Hu, Kyle Mahowald, Gary Lupyan, Anna Ivanova, and Roger Levy. 2024.
\newblock \href {https://doi.org/10.1073/pnas.2400917121} {Language models align with human judgments on key grammatical constructions}.
\newblock \emph{Proceedings of the National Academy of Sciences}, 121(36).

\bibitem[{Hu et~al.(2020)Hu, Ruder, Siddhant, Neubig, Firat, and Johnson}]{hu2020xtrememassivelymultilingualmultitask}
Junjie Hu, Sebastian Ruder, Aditya Siddhant, Graham Neubig, Orhan Firat, and Melvin Johnson. 2020.
\newblock \href {https://arxiv.org/abs/2003.11080} {Xtreme: A massively multilingual multi-task benchmark for evaluating cross-lingual generalization}.
\newblock \emph{Preprint}, arXiv:2003.11080.

\bibitem[{Ide et~al.(2025)Ide, Nishida, Vasselli, Oba, Sakai, Kamigaito, and Watanabe}]{ide-etal-2025-make}
Yusuke Ide, Yuto Nishida, Justin Vasselli, Miyu Oba, Yusuke Sakai, Hidetaka Kamigaito, and Taro Watanabe. 2025.
\newblock \href {https://doi.org/10.18653/v1/2025.naacl-long.380} {How to make the most of {LLM}s' grammatical knowledge for acceptability judgments}.
\newblock In \emph{Proceedings of the 2025 Conference of the Nations of the Americas Chapter of the Association for Computational Linguistics: Human Language Technologies (Volume 1: Long Papers)}, pages 7416--7432, Albuquerque, New Mexico. Association for Computational Linguistics.

\bibitem[{Jakub{\'i}{\v{c}}ek et~al.(2013)Jakub{\'i}{\v{c}}ek, Kilgarriff, Kov{\'a}{\v{r}}, Rychl{\'y}, and Suchomel}]{jakubicek2013tenten}
Milo{\v{s}} Jakub{\'i}{\v{c}}ek, Adam Kilgarriff, Vojt{\v{e}}ch Kov{\'a}{\v{r}}, Pavel Rychl{\'y}, and V{\'i}t Suchomel. 2013.
\newblock \href {https://www.sketchengine.eu/wp-content/uploads/The_TenTen_Corpus_2013.pdf} {The tenten corpus family}.
\newblock In \emph{Proceedings of the 7th International Corpus Linguistics Conference}, pages 125--127.

\bibitem[{Jakub{\'i}{\v{c}}ek et~al.(2010)Jakub{\'i}{\v{c}}ek, Kilgarriff, McCarthy, and Rychl{\'y}}]{jakubicek2010fast}
Milo{\v{s}} Jakub{\'i}{\v{c}}ek, Adam Kilgarriff, Diana McCarthy, and Pavel Rychl{\'y}. 2010.
\newblock \href {https://www.sketchengine.eu/wp-content/uploads/Fast_syntactic_2010.pdf} {Fast syntactic searching in very large corpora for many languages}.
\newblock \emph{PACLIC 24 Proceedings}, pages 741--747.

\bibitem[{Jeffreys(1946)}]{jeffreys1946invariant}
Harold Jeffreys. 1946.
\newblock \href {https://doi.org/10.1098/rspa.1946.0056} {An invariant form for the prior probability in estimation problems}.
\newblock \emph{Proceedings of the Royal Society of London. Series A. Mathematical and Physical Sciences}, 186(1007):453--461.

\bibitem[{Jiang et~al.(2020)Jiang, Xu, Araki, and Neubig}]{jiang-etal-2020-know}
Zhengbao Jiang, Frank~F. Xu, Jun Araki, and Graham Neubig. 2020.
\newblock \href {https://doi.org/10.1162/tacl_a_00324} {How can we know what language models know?}
\newblock \emph{Transactions of the Association for Computational Linguistics}, 8:423--438.

\bibitem[{Jiao et~al.(2024)Jiao, Liu, Tang, Matter, Pfeffer, and Anderson}]{jiao-etal-2024-spin}
Difan Jiao, Yilun Liu, Zhenwei Tang, Daniel Matter, J{\"u}rgen Pfeffer, and Ashton Anderson. 2024.
\newblock \href {https://doi.org/10.18653/v1/2024.findings-acl.277} {{SPIN}: Sparsifying and integrating internal neurons in large language models for text classification}.
\newblock In \emph{Findings of the Association for Computational Linguistics: ACL 2024}, pages 4666--4682, Bangkok, Thailand. Association for Computational Linguistics.

\bibitem[{Kilgarriff et~al.(2014)Kilgarriff, Baisa, Bušta, Jakubíček, Kovář, Michelfeit, Rychlý, and Suchomel}]{kilgarriff2014sketch}
Adam Kilgarriff, Vít Baisa, Jan Bušta, Miloš Jakubíček, Vojtěch Kovář, Jan Michelfeit, Pavel Rychlý, and Vít Suchomel. 2014.
\newblock \href {https://www.sketchengine.eu/wp-content/uploads/The_Sketch_Engine_2014.pdf} {The sketch engine: ten years on}.
\newblock \emph{Lexicography}, 1:7--36.

\bibitem[{Lai et~al.(2023)Lai, Ngo, Veyseh, Man, Dernoncourt, Bui, and Nguyen}]{lai2023chatgptenglishcomprehensiveevaluation}
Viet~Dac Lai, Nghia~Trung Ngo, Amir Pouran~Ben Veyseh, Hieu Man, Franck Dernoncourt, Trung Bui, and Thien~Huu Nguyen. 2023.
\newblock \href {https://arxiv.org/abs/2304.05613} {Chatgpt beyond english: Towards a comprehensive evaluation of large language models in multilingual learning}.
\newblock \emph{Preprint}, arXiv:2304.05613.

\bibitem[{Lin(1991)}]{lin1991divergence}
J.~Lin. 1991.
\newblock \href {https://doi.org/10.1109/18.61115} {Divergence measures based on the shannon entropy}.
\newblock \emph{IEEE Transactions on Information Theory}, 37(1):145--151.

\bibitem[{Liu et~al.(2019)Liu, Gardner, Belinkov, Peters, and Smith}]{liu2019linguistic}
Nelson~F. Liu, Matt Gardner, Yonatan Belinkov, Matthew~E. Peters, and Noah~A. Smith. 2019.
\newblock \href {https://doi.org/10.18653/v1/N19-1112} {Linguistic knowledge and transferability of contextual representations}.
\newblock In \emph{Proc. of NAACL-HLT}, pages 1073--1094, Minneapolis, Minnesota. Association for Computational Linguistics.

\bibitem[{Madabushi et~al.(2022)Madabushi, Gow-Smith, Garcia, Scarton, Idiart, and Villavicencio}]{madabushi2022semeval2022task2multilingual}
Harish~Tayyar Madabushi, Edward Gow-Smith, Marcos Garcia, Carolina Scarton, Marco Idiart, and Aline Villavicencio. 2022.
\newblock \href {https://arxiv.org/abs/2204.10050} {Semeval-2022 task 2: Multilingual idiomaticity detection and sentence embedding}.
\newblock \emph{Preprint}, arXiv:2204.10050.

\bibitem[{Malkiel(1959)}]{Malkiel1959StudiesII}
Yakov Malkiel. 1959.
\newblock \href {https://api.semanticscholar.org/CorpusID:53402704} {Studies in irreversible binomials}.
\newblock \emph{Lingua}, 8:113--160.

\bibitem[{Marks and Tegmark(2024)}]{marks2024geometrytruthemergentlinear}
Samuel Marks and Max Tegmark. 2024.
\newblock \href {https://arxiv.org/abs/2310.06824} {The geometry of truth: Emergent linear structure in large language model representations of true/false datasets}.
\newblock \emph{Preprint}, arXiv:2310.06824.

\bibitem[{{Meta}(2024)}]{meta2024llama32modelcard}
{Meta}. 2024.
\newblock Llama-3.2-3b model card.
\newblock \url{https://huggingface.co/meta-llama/Llama-3.2-3B}.

\bibitem[{Misra and Mahowald(2025)}]{misra2025languagemodelslearnrare}
Kanishka Misra and Kyle Mahowald. 2025.
\newblock \href {https://arxiv.org/abs/2403.19827} {Language models learn rare phenomena from less rare phenomena: The case of the missing aanns}.
\newblock \emph{Preprint}, arXiv:2403.19827.

\bibitem[{Mollin(2012)}]{mollin2012}
Sandra Mollin. 2012.
\newblock \href {https://doi.org/10.1017/S1360674311000293} {Revisiting binomial order in english: ordering constraints and reversibility}.
\newblock \emph{English Language and Linguistics}, 16(1):81–103.

\bibitem[{Morgan and Levy(2015)}]{morgan2015modeling}
Emily Morgan and Roger Levy. 2015.
\newblock \href {https://escholarship.org/uc/item/7rm1r6jh} {Modeling idiosyncratic preferences: How generative knowledge and expression frequency jointly determine language structure}.
\newblock In \emph{Proceedings of the Annual Meeting of the Cognitive Science Society}, volume~37.

\bibitem[{Morgan and Levy(2016)}]{morgan2016abstract}
Emily Morgan and Roger Levy. 2016.
\newblock \href {https://doi.org/10.1016/j.cognition.2016.09.011} {Abstract knowledge versus direct experience in processing of binomial expressions}.
\newblock \emph{Cognition}, 157:384--402.

\bibitem[{Motschenbacher(2013)}]{motschenbacher2013gentlemen}
Heiko Motschenbacher. 2013.
\newblock \href {https://doi.org/10.1177/0075424213489993} {Gentlemen before ladies? a corpus-based study of conjunct order in personal binomials}.
\newblock \emph{Journal of English Linguistics}, 41(3):212--242.

\bibitem[{Panickssery et~al.(2024)Panickssery, Gabrieli, Schulz, Tong, Hubinger, and Turner}]{panickssery2024steeringllama2contrastive}
Nina Panickssery, Nick Gabrieli, Julian Schulz, Meg Tong, Evan Hubinger, and Alexander~Matt Turner. 2024.
\newblock \href {https://arxiv.org/abs/2312.06681} {Steering llama 2 via contrastive activation addition}.
\newblock \emph{Preprint}, arXiv:2312.06681.

\bibitem[{Pinker and Birdsong(1979)}]{pinker1979speakers}
Steven Pinker and David Birdsong. 1979.
\newblock \href {https://doi.org/10.1016/S0022-5371(79)90273-1} {Speakers' sensitivity to rules of frozen word order}.
\newblock \emph{Journal of Verbal Learning and Verbal Behavior}, 18(4):497--508.

\bibitem[{Qiu et~al.(2024)Qiu, Duan, and Cai}]{qiu-etal-2024-evaluating}
Zhuang Qiu, Xufeng Duan, and Zhenguang Cai. 2024.
\newblock \href {https://doi.org/10.18653/v1/2024.cmcl-1.16} {Evaluating grammatical well-formedness in large language models: A comparative study with human judgments}.
\newblock In \emph{Proceedings of the Workshop on Cognitive Modeling and Computational Linguistics}, pages 189--198, Bangkok, Thailand. Association for Computational Linguistics.

\bibitem[{Radford et~al.(2018)Radford, Narasimhan, Salimans, Sutskever et~al.}]{radford2018improving}
Alec Radford, Karthik Narasimhan, Tim Salimans, Ilya Sutskever, and 1 others. 2018.
\newblock \href {https://www.mikecaptain.com/resources/pdf/GPT-1.pdf} {Improving language understanding by generative pre-training}.

\bibitem[{Ravfogel et~al.(2020)Ravfogel, Elazar, Gonen, Twiton, and Goldberg}]{ravfogel-etal-2020-null}
Shauli Ravfogel, Yanai Elazar, Hila Gonen, Michael Twiton, and Yoav Goldberg. 2020.
\newblock \href {https://doi.org/10.18653/v1/2020.acl-main.647} {Null it out: Guarding protected attributes by iterative nullspace projection}.
\newblock In \emph{Proceedings of the 58th Annual Meeting of the Association for Computational Linguistics}, pages 7237--7256, Online. Association for Computational Linguistics.

\bibitem[{Reinhart et~al.(2025)Reinhart, Markey, Laudenbach, Pantusen, Yurko, Weinberg, and Brown}]{reinhart2025llms}
Alex Reinhart, Benjamin Markey, Matt Laudenbach, Kaitlyn Pantusen, Ronald Yurko, Graham Weinberg, and David~W. Brown. 2025.
\newblock \href {https://doi.org/10.1073/pnas.2422455122} {Do llms write like humans? variation in grammatical and rhetorical styles}.
\newblock \emph{Proceedings of the National Academy of Sciences}, 122(8):e2422455122.

\bibitem[{Spearman(1904)}]{spearman1904proof}
Charles Spearman. 1904.
\newblock \href {https://doi.org/10.2307/1412159} {The proof and measurement of association between two things}.
\newblock \emph{The American Journal of Psychology}, 15(1):72--101.

\bibitem[{Team et~al.(2025)Team, Kamath, Ferret, Pathak, Vieillard, Merhej, Perrin, Matejovicova, Ramé, Rivière, Rouillard, Mesnard, Cideron, bastien Grill, Ramos, Yvinec, Casbon, Pot, Penchev, Liu, Visin, Kenealy, Beyer, Zhai, Tsitsulin, Busa-Fekete, Feng, Sachdeva, Coleman, Gao, Mustafa, Barr, Parisotto, Tian, Eyal, Cherry, Peter, Sinopalnikov, Bhupatiraju, Agarwal, Kazemi, Malkin, Kumar, Vilar, Brusilovsky, Luo, Steiner, Friesen, Sharma, Sharma, Gilady, Goedeckemeyer, Saade, Feng, Kolesnikov, Bendebury, Abdagic, Vadi, György, Pinto, Das, Bapna, Miech, Yang, Paterson, Shenoy, Chakrabarti, Piot, Wu, Shahriari, Petrini, Chen, Lan, Choquette-Choo, Carey, Brick, Deutsch, Eisenbud, Cattle, Cheng, Paparas, Sreepathihalli, Reid, Tran, Zelle, Noland, Huizenga, Kharitonov, Liu, Amirkhanyan, Cameron, Hashemi, Klimczak-Plucińska, Singh, Mehta, Lehri, Hazimeh, Ballantyne, Szpektor, Nardini, Pouget-Abadie, Chan, Stanton, Wieting, Lai, Orbay, Fernandez, Newlan, yeong Ji, Singh, Black, Yu, Hui, Vodrahalli, Greff, Qiu,
  Valentine, Coelho, Ritter, Hoffman, Watson, Chaturvedi, Moynihan, Ma, Babar, Noy, Byrd, Roy, Momchev, Chauhan, Sachdeva, Bunyan, Botarda, Caron, Rubenstein, Culliton, Schmid, Sessa, Xu, Stanczyk, Tafti, Shivanna, Wu, Pan, Rokni, Willoughby, Vallu, Mullins, Jerome, Smoot, Girgin, Iqbal, Reddy, Sheth, Põder, Bhatnagar, Panyam, Eiger, Zhang, Liu, Yacovone, Liechty, Kalra, Evci, Misra, Roseberry, Feinberg, Kolesnikov, Han, Kwon, Chen, Chow, Zhu, Wei, Egyed, Cotruta, Giang, Kirk, Rao, Black, Babar, Lo, Moreira, Martins, Sanseviero, Gonzalez, Gleicher, Warkentin, Mirrokni, Senter, Collins, Barral, Ghahramani, Hadsell, Matias, Sculley, Petrov, Fiedel, Shazeer, Vinyals, Dean, Hassabis, Kavukcuoglu, Farabet, Buchatskaya, Alayrac, Anil, Dmitry, Lepikhin, Borgeaud, Bachem, Joulin, Andreev, Hardin, Dadashi, and Hussenot}]{gemmateam2025gemma3technicalreport}
Gemma Team, Aishwarya Kamath, Johan Ferret, Shreya Pathak, Nino Vieillard, Ramona Merhej, Sarah Perrin, Tatiana Matejovicova, Alexandre Ramé, Morgane Rivière, Louis Rouillard, Thomas Mesnard, Geoffrey Cideron, Jean bastien Grill, Sabela Ramos, Edouard Yvinec, Michelle Casbon, Etienne Pot, Ivo Penchev, and 197 others. 2025.
\newblock \href {https://arxiv.org/abs/2503.19786} {Gemma 3 technical report}.
\newblock \emph{Preprint}, arXiv:2503.19786.

\bibitem[{Tenney et~al.(2019)Tenney, Das, and Pavlick}]{tenney2019bertrediscoversclassicalnlp}
Ian Tenney, Dipanjan Das, and Ellie Pavlick. 2019.
\newblock \href {https://arxiv.org/abs/1905.05950} {Bert rediscovers the classical nlp pipeline}.
\newblock \emph{Preprint}, arXiv:1905.05950.

\bibitem[{Tibshirani(1996)}]{tibshirani1996regression}
Robert Tibshirani. 1996.
\newblock \href {https://academic.oup.com/jrsssb/article/58/1/267/7027929} {Regression shrinkage and selection via the lasso}.
\newblock \emph{Journal of the Royal Statistical Society Series B: Statistical Methodology}, 58(1):267--288.

\bibitem[{Tigges et~al.(2023)Tigges, Hollinsworth, Geiger, and Nanda}]{tigges2023linear}
Curt Tigges, Oskar~John Hollinsworth, Atticus Geiger, and Neel Nanda. 2023.
\newblock \href {https://arxiv.org/abs/2310.15154} {Linear representations of sentiment in large language models}.
\newblock \emph{ArXiv preprint}, abs/2310.15154.

\bibitem[{Turner et~al.(2023)Turner, Thiergart, Udell, Leech, Mini, and MacDiarmid}]{turner2023activation}
Alex Turner, Lisa Thiergart, David Udell, Gavin Leech, Ulisse Mini, and Monte MacDiarmid. 2023.
\newblock \href {https://arxiv.org/abs/2308.10248} {Activation addition: Steering language models without optimization}.
\newblock \emph{ArXiv preprint}, abs/2308.10248.

\bibitem[{Turner et~al.(2024)Turner, Thiergart, Leech, Udell, Vazquez, Mini, and MacDiarmid}]{turner2024steeringlanguagemodelsactivation}
Alexander~Matt Turner, Lisa Thiergart, Gavin Leech, David Udell, Juan~J. Vazquez, Ulisse Mini, and Monte MacDiarmid. 2024.
\newblock \href {https://arxiv.org/abs/2308.10248} {Steering language models with activation engineering}.
\newblock \emph{Preprint}, arXiv:2308.10248.

\bibitem[{Yang et~al.(2025)Yang, Li, Yang, Zhang, Hui, Zheng, Yu, Gao, Huang, Lv, Zheng, Liu, Zhou, Huang, Hu, Ge, Wei, Lin, Tang, Yang, Tu, Zhang, Yang, Yang, Zhou, Zhou, Lin, Dang, Bao, Yang, Yu, Deng, Li, Xue, Li, Zhang, Wang, Zhu, Men, Gao, Liu, Luo, Li, Tang, Yin, Ren, Wang, Zhang, Ren, Fan, Su, Zhang, Zhang, Wan, Liu, Wang, Cui, Zhang, Zhou, and Qiu}]{yang2025qwen3technicalreport}
An~Yang, Anfeng Li, Baosong Yang, Beichen Zhang, Binyuan Hui, Bo~Zheng, Bowen Yu, Chang Gao, Chengen Huang, Chenxu Lv, Chujie Zheng, Dayiheng Liu, Fan Zhou, Fei Huang, Feng Hu, Hao Ge, Haoran Wei, Huan Lin, Jialong Tang, and 41 others. 2025.
\newblock \href {https://arxiv.org/abs/2505.09388} {Qwen3 technical report}.
\newblock \emph{Preprint}, arXiv:2505.09388.

\bibitem[{Zou et~al.(2025)Zou, Phan, Chen, Campbell, Guo, Ren, Pan, Yin, Mazeika, Dombrowski, Goel, Li, Byun, Wang, Mallen, Basart, Koyejo, Song, Fredrikson, Kolter, and Hendrycks}]{zou2025representationengineeringtopdownapproach}
Andy Zou, Long Phan, Sarah Chen, James Campbell, Phillip Guo, Richard Ren, Alexander Pan, Xuwang Yin, Mantas Mazeika, Ann-Kathrin Dombrowski, Shashwat Goel, Nathaniel Li, Michael~J. Byun, Zifan Wang, Alex Mallen, Steven Basart, Sanmi Koyejo, Dawn Song, Matt Fredrikson, and 2 others. 2025.
\newblock \href {https://arxiv.org/abs/2310.01405} {Representation engineering: A top-down approach to ai transparency}.
\newblock \emph{Preprint}, arXiv:2310.01405.

\end{thebibliography}
